\pdfoutput=1
\documentclass[11pt]{article}

\usepackage[]{ACL}

\usepackage{times}
\usepackage{latexsym}

\usepackage[T1]{fontenc}
\usepackage[utf8]{inputenc}

\usepackage{microtype}

\usepackage{inconsolata}

\usepackage{multicol}
\usepackage{multirow}
\usepackage{amssymb}
\usepackage{amsmath}
\usepackage{xcolor}
\usepackage{calc}
\usepackage{amsfonts}
\usepackage{graphicx}
\usepackage{tabularx}
\usepackage{booktabs}
\usepackage{CJKutf8}
\usepackage{makecell}
\usepackage{color}
\usepackage{xcolor}
\usepackage{colortbl} 
\usepackage[fixed]{fontawesome5}
\usepackage[ruled,vlined]{algorithm2e}
\usepackage{tcolorbox}
\usepackage{algpseudocode}
\newcommand{\tcpo}[1]{\tcp{\textrm{\textcolor{blue}{#1}}}}

\definecolor{babyblue}{RGB}{217, 239, 251}

\definecolor{light_blue}{HTML}{6699CC}
\definecolor{medium_blue}{HTML}{336699}
\definecolor{dark_blue}{HTML}{003366}
\definecolor{lightyellow}{RGB}{252, 245, 177}
\definecolor{mellowyellow}{RGB}{254, 245, 218}
\definecolor{lightgray}{gray}{0.9}
\definecolor{darkblue_fig}{RGB}{37,95,157}
\definecolor{green_fig}{RGB}{21,165,63}
\definecolor{red_fig}{RGB}{176,0,10}
\definecolor{lightgrey_fig}{RGB}{108,108,108}
\definecolor{lightorange}{HTML}{F39652}

\newtcbox{\hlprimarytab}{on line, rounded corners, box align=base, colback=c3!10,colframe=white,size=fbox,arc=3pt, before upper=\strut, top=-2pt, bottom=-4pt, left=-2pt, right=-2pt, boxrule=0pt}
\newtcbox{\hlsecondarytab}{on line, box align=base, colback=orange!20,colframe=white,size=fbox,arc=3pt, before upper=\strut, top=-2pt, bottom=-4pt, left=-2pt, right=-2pt, boxrule=0pt}
\newcommand{\daugshifted}{\raisebox{0.5\depth}{$\uparrow$}}

\newcommand{\daulg}[1]{{\hlsecondarytab{\daugshifted{#1}}}}

\usepackage{graphicx,calc}
\usepackage{wrapfig}
\newlength\myheight
\newlength\mydepth
\settototalheight\myheight{Xygp}
\newcommand{\ours}{\textsc{KnowAgent}}

\title{KnowAgent: Knowledge-Augmented Planning for LLM-Based Agents}

\author{Yuqi Zhu\textsuperscript{$\clubsuit$$\heartsuit$},
Shuofei Qiao\textsuperscript{$\clubsuit$$\heartsuit$},
Yixin Ou\textsuperscript{$\clubsuit$$\heartsuit$},
Shumin Deng\textsuperscript{$\spadesuit$},
{\bf Shiwei Lyu}$^{\diamondsuit}$,
\\
{\bf Yue Shen}$^{\diamondsuit}$,
{\bf Lei Liang}$^{\diamondsuit}$,
{\bf Jinjie Gu}$^{\diamondsuit}$,
\textbf{Huajun Chen}\textsuperscript{$\clubsuit$$\heartsuit$},
\textbf{Ningyu Zhang}\textsuperscript{$\clubsuit$$\heartsuit$}\thanks{~~Corresponding author.}
\\
$^\clubsuit$ Zhejiang University
 $^\heartsuit$ ZJU-Ant Group Joint Research Center for Knowledge Graphs\\
$^\diamondsuit$ Ant Group 
$^\spadesuit$ National University of Singapore, NUS-NCS Joint Lab, Singapore
\\
 \texttt{\{zhuyuqi,zhangningyu\}@zju.edu.cn}\\
  \raisebox{-\mydepth}{\includegraphics[height=1.6\myheight]{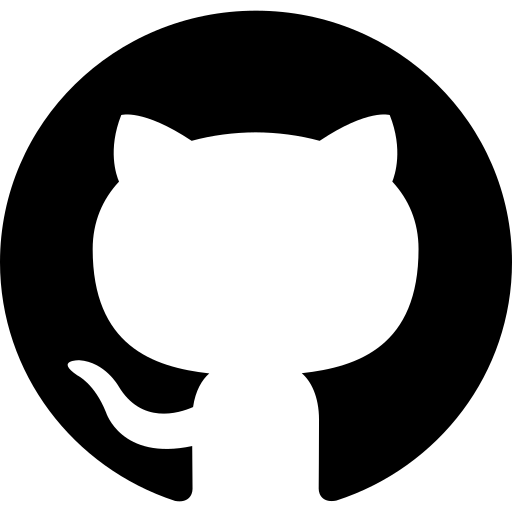}}
\textbf{\url{https://zjunlp.github.io/project/KnowAgent/}}
}

\begin{document}
\maketitle
\begin{abstract}
Large Language Models (LLMs) have demonstrated great potential in complex reasoning tasks, yet they fall short when tackling more sophisticated challenges, especially when interacting with environments through generating executable actions. This inadequacy primarily stems from the lack of built-in action knowledge in language agents, which fails to effectively guide the planning trajectories during task solving and results in \textit{planning hallucination}. To address this issue, we introduce \textbf{\ours}, a novel approach designed to enhance the planning capabilities of LLMs by incorporating explicit action knowledge. Specifically, \textbf{\ours} employs an \textit{action knowledge base} and a \textit{knowledgeable self-learning} strategy to constrain the action path during planning, enabling more reasonable trajectory synthesis, and thereby enhancing the planning performance of language agents. Experimental results on HotpotQA and ALFWorld based on various backbone models demonstrate that \textbf{\ours} can achieve comparable or superior performance to existing baselines. Further analysis indicates the effectiveness of \textbf{\ours} in terms of \textit{planning hallucinations} mitigation.
% \footnote{Code is at \url{https://github.com/zjunlp/KnowAgent}.}.
\end{abstract}

\section{Introduction}
% As artificial intelligence (AI) advances,
% language agents have emerged as pivotal tools for tackling complex problems. 
As artificial intelligence (AI) advances, language agents are becoming increasingly vital for solving complex problems~\citep{sjtu/cot/survey,sumers2024cognitive,code/agent/survey}. 
These agents, built around Large Language Models (LLMs), enhance their task planning capabilities through a variety of strategies including task decomposition~\citep{cot,tot,voyager,xagent}, reflection
~\citep{reflexion,adaplanner}, 
% ~\citep{reflexion,self-refine,adaplanner}, 
collaborative division of labor
~\citep{metagpt,agentverse,lumos,autoact}, and the utilization of external tools~\citep{toolformer,trice}.
Despite the effectiveness of current prompting techniques in providing good planning abilities for some closed-source language models, these methods are often limited by the model's intrinsic understanding capabilities and the scope of knowledge it was trained on. 
To meet the demands for broad application and customization in different areas such as question-answering~\citep{retroformer,lumos}, web browsing~\citep{webshop,mind2web,webarena}, robotics~\citep{Do_As_I_Can,actionknowledge/robots} and so on, researchers are exploring \textit{Agent Tuning} as a means to augment model capabilities~\citep{fireact,agenttuning,kwaiagents}. 
% This involves fine-tuning models through the synthesis of task-specific trajectories, enabling them to undertake a series of effective actions to complete tasks, thereby enhancing their ability to handle complex situations.
This involves fine-tuning models through the synthesis of task-specific trajectories, enabling them to undertake a series of effective actions to handle complex situations.

\begin{figure}[!t]
\centering 
\includegraphics[width=0.45\textwidth]{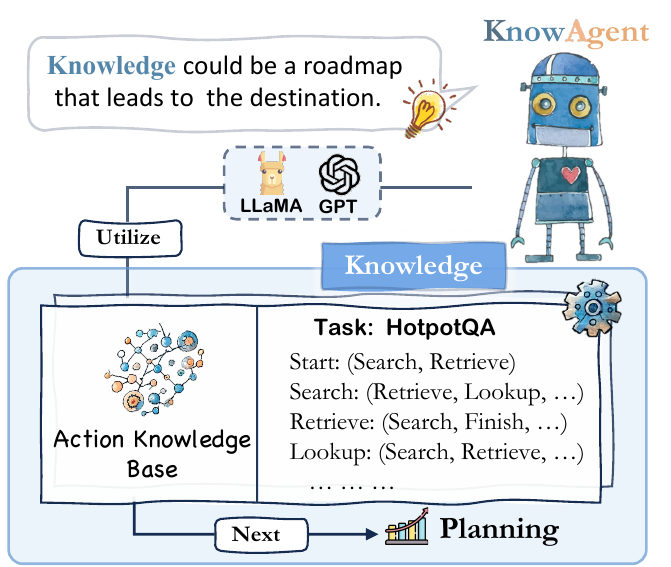} 
\caption{\textbf{The overview of {\ours}}. An agent could leverage external action knowledge base to address and solve complex planning challenges.}
\label{fig:overall}
\end{figure}

However, when it comes to executing planning tasks, especially in open-source models, there remain issues~\citep{agentbench,llm/self-critic/plans,llm/world_model/planning}.
Frequently, models generate plans that violate established knowledge rules or commonsense~\citep{actionknowledge/robots}, a phenomenon we name as \textit{planning hallucination}. 
This term describes scenarios where models might generate unnecessary or conflicting action sequences, such as \texttt{``attempting to look up information without performing a search operation''} or \texttt{``trying to pick an apple from a table without verifying the presence of both the table and the apple''}. 
%These inaccuracies not only diminish the effectiveness of the models but also constrain the further development of AI technologies. 

% In response to these challenges, we introduce an innovative framework: {\ours}. 
To address these issues, we propose \textbf{\ours} that focuses on leveraging external \textit{action knowledge} to enhance synthetic trajectories with the goal of resolving \textit{planning hallucination} (see Figure~\ref{fig:overall}).
Our development is grounded on several key steps: Initially, we create an extensive \textit{action knowledge base}, which amalgamates action planning knowledge pertinent to specific tasks. 
This database acts as an external reservoir of information, steering the model's action generation process. 
Subsequently, by converting \textit{action knowledge} into text, we enable the model to deeply understand and utilize this knowledge in creating action trajectories. 
Finally, through a \textit{knowledgeable self-learning} phase, we use trajectories developed from the model's iterative processes to continually improve its understanding and application of \textit{action knowledge}. 
This process not only strengthens the agents' planning abilities but also enhances their potential for application in complex situations.

Experimental results on HotpotQA~\citep{datasets/hotpotqa} and ALFWorld~\citep{datasets/alfworld} based on various backbone models demonstrate that \textbf{\ours} can achieve comparable or superior performance to existing baselines.
Further analysis indicates the effectiveness of \textbf{\ours} in terms of \textit{planning hallucinations} mitigation.
We summarize our contributions as follows:
\begin{itemize}
    \item We introduce {\ours} that employs \textit{knowledgeable self-learning} to incorporate external \textit{action knowledge} into models to refine and augment the intrinsic planning abilities of language agents.
    \item We conduct comprehensive experiments that demonstrate  {\ours} can match or surpass other benchmark models on the HotpotQA and ALFWorld datasets. 
    \item Further analysis validates the effectiveness of incorporating \textit{action knowledge} for planning purposes. 
    We also showcase the possibility of employing manually refined \textit{action knowledge} from LLMs, thereby reducing human labor and enhancing performance.
\end{itemize}

\section{Background}
\label{background}

Language agents observe the external world primarily by generating inner thoughts and executable actions.
In this paper, we follow and further enhance the planning trajectory format proposed in \citet{react}
to train and evaluate our \ours.
Traditionally, a planning trajectory $\tau$ can be represented by a triplet of Thought-Action-Observation $(\mathcal{T},\mathcal{A},\mathcal{O})$, where $\mathcal{T}$ indicates the inner thoughts of the language agent, $\mathcal{A}$ signifies executable actions, and $\mathcal{O}$ represents the feedback information from the environment.
In terms of this, the trajectory history $\mathcal{H}$ at time $t$ can be defined as:
$
% \begin{align}
    \mathcal{H}_t=(\mathcal{T}_0,\mathcal{A}_0,\mathcal{O}_0, \mathcal{T}_1, ..., \mathcal{T}_{t-1},\mathcal{A}_{t-1},\mathcal{O}_{t-1}).
% \end{align}
$
Then, the language agent is reinforced to generate $\mathcal{T}_t$ and $\mathcal{A}_t$ based on the history.
Given a parameterized probabilistic language agent $\pi$ with parameters $\theta$, the process of generating the next step's thought based on $\mathcal{H}_t$ can be represented as:
\begin{align}
    \label{eq:T_generate}
    p(\mathcal{T}_t|\mathcal{H}_t) = \prod_{i=1}^{|\mathcal{T}_t|} \pi_\theta(\mathcal{T}_t^{i}|\mathcal{H}_t,\mathcal{T}_t^{<i}),
\end{align}
where $\mathcal{T}_t^{i}$ and $|\mathcal{T}_t|$ are the $i$-th token and the length of $\mathcal{T}_t$ respectively.
Subsequently, the action $\mathcal{A}_t$ will be determined based on $\mathcal{T}_t$ and $\mathcal{H}_t$:
\begin{align}
    \label{eq:A_generate}
    p(\mathcal{A}_t|\mathcal{H}_t,\mathcal{T}_t) = \prod_{j=1}^{|\mathcal{A}_t|} \pi_\theta(\mathcal{A}_t^{j}|\mathcal{H}_t,\mathcal{T}_t,\mathcal{A}_t^{<j}).
\end{align}
Similarly, $\mathcal{A}_t^{j}$ and $|\mathcal{A}_t|$ denote the $j$-th token and the length of $\mathcal{A}_t$ respectively.
Lastly, the feedback result of the action $\mathcal{A}_t$ will be treated as the observation $\mathcal{O}_t$ and added to the trajectory, generating a new round of trajectory $\mathcal{H}_{t+1}$.
It's important to note that $\mathcal{A}_i$ here specifically means the actions in the trajectory, which \textbf{is identical to} the action $a_i$ in the discussion of the action set $E_a$ later on.

\begin{figure*}[!t]

\centering 
\includegraphics[width=.94\textwidth]{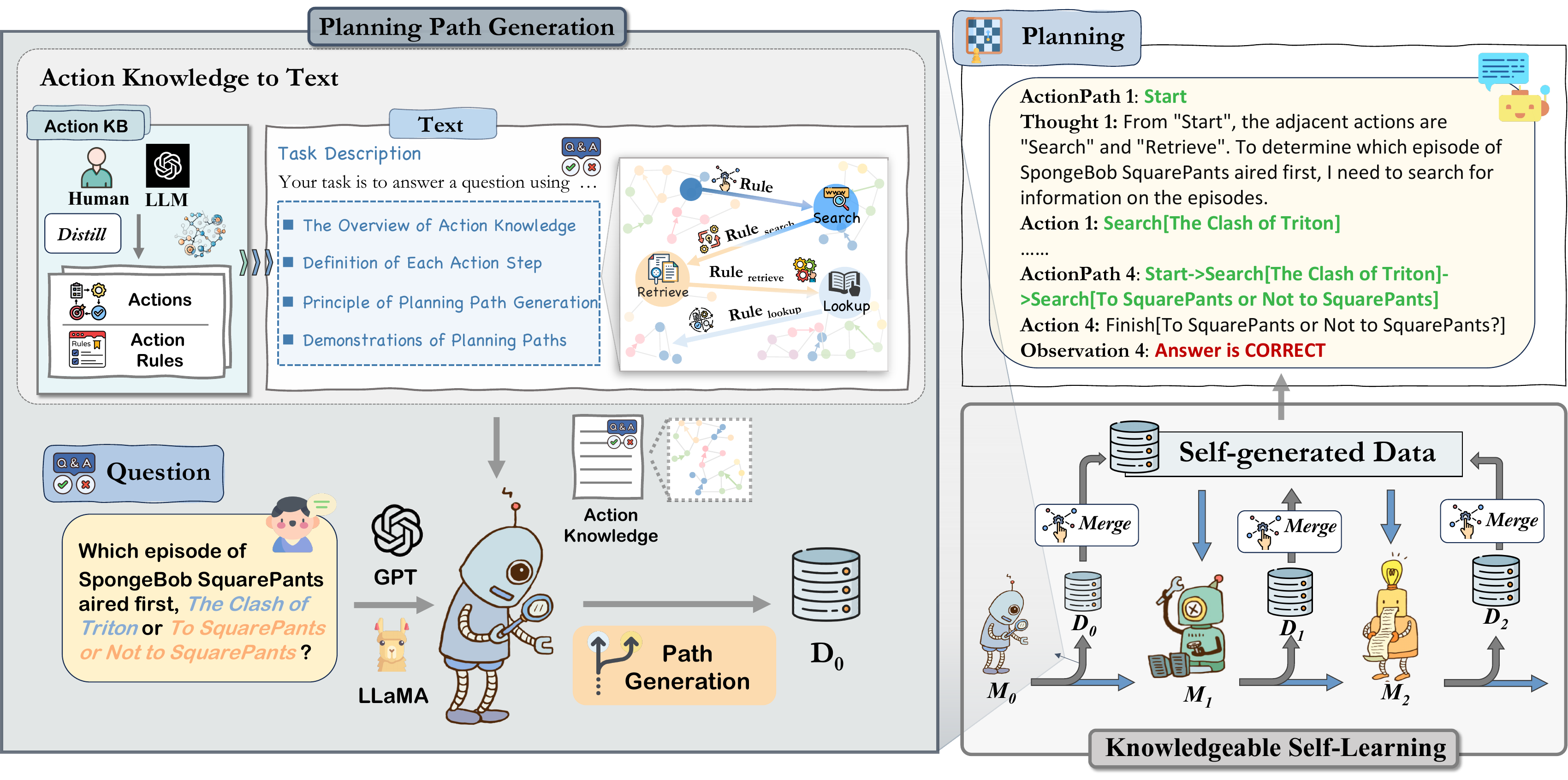} 
\caption{\textbf{The overall framework of {\ours}}. 
Initially, \textit{Action Knowledge to Text} converts task-specific action knowledge into textual descriptions. 
Next, \textit{Planning Path Generation} uses prompts and this knowledge to lead LLMs in planning path creation. 
Lastly, in \textit{Knowledgeable Self-Learning},  the model iteratively optimizes using generated planning trajectories to improve performance.
 % Finally, the \textit{Self-Learning} phase employs the planning paths and their resulting trajectory sequences generated by the model for iterative optimization, aiming to enhance model performance.
 }
 \vskip -0.05in
\label{fig:framework}
\end{figure*}

\section{\ours}
Our method, {\ours}, as illustrated in Figure~\ref{fig:framework}, begins by defining \textit{action knowledge}. 
It then utilizes this knowledge to generate planning paths, and continuously refines these paths through a \textit{knowledgeable self-learning} mechanism, enhancing the framework iteratively.

\subsection{Definition of Action Knowledge}
\label{sec:define_action_knowledge}
\paragraph{Action.}
% The symbol $\mathcal{A}$ signifies the set of Actions, which encompasses the discrete action that LLMs must undertake to accomplish specific tasks.
% The symbol $\mathcal{A}$ denotes a collection of actions, comprising discrete steps that LLMs are required to perform to achieve specific tasks. 
% Within this set, each $a_i \in \mathcal{A}$ specifies a distinct action. 
% For any specified task, a subset $\mathcal{A}_t \subseteq \mathcal{A}$ is identified for implementation.
$E_a = \{a_1,...,a_{N-1}\}$ signifies a set of actions, which encompasses the discrete action that LLMs must undertake to accomplish specific tasks.

\paragraph{Action Rules.}
% Action rules $\mathcal{R}$ establish the logic and sequence for transitions between nodes in the action knowledge, clarifying which actions may proceed and the sequence of execution based on predefined conditions. 
% For example, a rule may stipulate that a transition can occur from an action $E_a$ to node $\mathcal{N}_j$ via action $a_j$, or to node $\mathcal{N}_k$ via action $a_k$, which is formally represented as ``
% $\mathcal{N}_i \xrightarrow{a_j} \mathcal{N}_j$, $\mathcal{N}_i \xrightarrow{a_k} \mathcal{N}_k$
% '', respectively.
% These rules are valuable in complex situations, providing guidance on possible state transitions and aiding in the prediction of pathways from one to another.
$\mathcal{R} = \{r_1,...,r_{N-1}\}$ outlines the rules that determine the logic and sequence of action transitions within the model. 
These rules directly dictate permissible action transitions $r_k : a_i \to a_j$, based on the inherent relationships among actions or task-specific requirements.

\paragraph{Action Knowledge.}
Action Knowledge, represented as $(E_a, \mathcal{R})$, comprises a defined set of actions $E_a$ and the rules $\mathcal{R}$ governing their transitions. 
The combination of \textit{action knowledge} for different tasks forms an \textit{action knowledge base}, also known as \textit{\textbf{Action KB}}. 
The knowledge base will then serve as essential guidance for generating actions and formulating decisions, essential for reducing potential \textit{plan hallucination} problems.
% About Action Knowledge, which is denoted by $\mathcal{AK}$, it encapsulates a well-defined ensemble of action sets along with their corresponding rules and is formalized as $\mathcal{AK = \{A, R\}}$. 
% Here $\mathcal{A}$ signifies the collection of possible actions pertinent to a specific domain or context, while $\mathcal{R}$ encompasses the set of rules that govern the logic and sequence of these actions.

\paragraph{Strategies for Extracting Action Knowledge.}

Given the diverse \textit{action knowledge} involved in various tasks, fully manual construction is both time-consuming and labor-intensive. 
To overcome this, we leverage GPT-4~\citep{gpt4}, known for its strong performance on such tasks~\citep{agentbench,autoplan}, for initial construction. 
Our two-stage process starts with domain experts providing task knowledge to the LLM, which generates a preliminary list of actions and rules. 
Since the initial output often includes redundancy, human experts then filter and refine this list. 
In the second stage, the refined actions and specifications are reintroduced to the LLM to produce the final set of action rules.
% In \S~\ref{sec:analysis}, we provide a detailed comparison of the effectiveness of these two approaches.
A detailed comparison of these two approaches is provided in \S~\ref{sec:analysis}.

\subsection{Planning Path Generation with Action Knowledge}
\label{sec:knowledge2path}
\subsubsection{Action Knowledge to Text}
\label{sec:ak2text}
% Upon acquiring action knowledge pertinent to a specific task, we transform it into a textual format to facilitate a deeper understanding by the model.
% Specifically, the action knowledge in the extraction process is illustrated in Figure \ref{fig:prompt}, exemplified by the notation \textit{Search:(Search, Retrieve, Lookup, Finish)}.
% This notation indicates that at the \textit{Search} node, there are multiple action pathways: the action can either recur as \textit{Search}, transition to \textit{Retrieve} or \textit{Lookup}, or progress towards \textit{Finish}. 
Figure~\ref{fig:framework} illustrates the conversion process from \textit{action knowledge} to text. 
Initially, we establish the \textit{action knowledge base} by identifying actions pertinent to the task's specific needs, utilizing previous dataset analyses and the inherent knowledge of LLMs.
% This information is then presented in text format to facilitate a more profound understanding by the model.
This information is then converted into text format to facilitate subsequent operations.
As an illustration, we reference one action rule in HotpotQA~\citep{datasets/hotpotqa} - \texttt{Search:(Search, Retrieve, Lookup, Finish)}. 
% This rule signifies that from \texttt{Search}, several pathways are viable: an action may continue as \texttt{Search}, evolve into \texttt{Retrieve} or \texttt{Lookup}, or advance towards \texttt{Finish}.
This rule indicates that \texttt{Search} can lead to multiple pathways: continuing as \texttt{Search}, changing to \texttt{Retrieve} or \texttt{Lookup}, or progressing to \texttt{Finish}.

\subsubsection{Path Generation}
\label{sec:pathgeneration}
Harnessing \textit{action knowledge}, the model utilizes this insight to streamline the task's planning process.
It achieves this by formulating a coherent planning path, guided by the application of action rules $\mathcal{R}_1 \land \mathcal{R}_2 \land \ldots \Rightarrow \mathcal{P}$. 
To facilitate path generation, we develop specialized prompts that extend beyond basic \textbf{Task Description}, integrating segments 
as illustrated in Figure \ref{fig:framework} (or in Figure \ref{fig:prompt}).
% as illustrated in Figure \ref{fig:framework}~(detailed in Figure \ref{fig:prompt}).
% In addition to the basic \texttt{Task Description}, the methodology incorporates several crucial elements:

Our approach is thoroughly grounded in \textit{action knowledge} and unfolds across four key segments:
(1) It starts with an \textbf{Overview of Action Knowledge} to set the foundational concepts and rules.  
(2) This is followed by the \textbf{Definition of Each Action Step}, detailing the operational aspects and significance of each action.
(3) Following this, the \textbf{Principle of Planning Path Generation} delves into the constraints on output generation.  
(4) And finally, \textbf{Demonstrations of Planning Paths} provide practical examples, acting as a beacon of inspiration for adapting these strategies across various contexts.
Each of these segments plays an essential role in expressing \textit{action knowledge}, specifying actions, and clarifying the process of leveraging \textit{action knowledge} for planning path generation.
% Each of these segments plays an essential role in clarifying the process of leveraging action knowledge for planning path generation.
\textbf{It's essential to understand the distinction between path and trajectory in this context}.
The \textbf{path} exclusively represents the series of actions undertaken by the agent, while the \textbf{trajectory} includes the model's complete output during the problem-solving process, incorporating the path as part of its structure.

Here we briefly outline the process of \textbf{trajectory} synthesis.
This trajectory, denoted as $\tau$, is composed of many planned quadruples.
Each quadruple $(\mathcal{P},\mathcal{T},\mathcal{A},\mathcal{O})$, encapsulates the action path $\mathcal{P}$, the agent’s internal thoughts processes $\mathcal{T}$, executable actions $\mathcal{A}$, and environmental feedback $\mathcal{O}$.
The historical trajectory is reformulated as:
% \begin{gather}
$
    \mathcal{H}_t = \nonumber 
    (\mathcal{P}_0, \mathcal{T}_0, \mathcal{A}_0, \mathcal{O}_0,\ldots, \mathcal{T}_{t-1}, \mathcal{A}_{t-1}, \mathcal{O}_{t-1}).
$
% \end{gather}
% The model's trajectory $\tau$ is composed of numerous planned quadruples. Each quadruple $(p, t, a, o)$ encapsulates the action path $p$, the agent's internal thought process $t$, executable actions $a$, and feedback from the environment $o$.
Based on this, 
% historical trajectory
 the agent is poised to generate $\mathcal{P}_t$, $\mathcal{T}_t$, and $\mathcal{A}_t$. 
% We consider a parameterized probabilistic language model, denoted as $\pi$ with parameters $\theta$. 
% The generation of the action path conditional on $\mathcal{P}_t$, is defined by:
Considering a parameterized probabilistic language agent $\pi$ with parameters $\theta$, the mechanism for generating the subsequent action path, contingent on $\mathcal{P}_t$, is expressed as:
% \begin{align}
$
    p(\mathcal{P}_t|\mathcal{H}_t) = \prod_{k=1}^{|\mathcal{P}_t|} \pi_\theta(\mathcal{P}_t^{k}|\mathcal{H}_t,\mathcal{P}_t^{<k}).
% \end{align}
$
Here $\mathcal{P}_t^{k}$ and $|\mathcal{P}_t|$ represent the $k$-th token and the total length of $\mathcal{P}_t$. 
And then we extend the approach used in 
Equation \ref{eq:T_generate} and \ref{eq:A_generate}.
The process of deriving thoughts and actions can be reformulated as:
\begin{align}
p(\mathcal{T}_t|\mathcal{H}_t,\mathcal{P}_t) &= \prod_{i=1}^{|\mathcal{T}_t|} \pi_\theta(\mathcal{T}_t^{i}|\mathcal{H}_t,\mathcal{P}_t,\mathcal{T}_t^{<i}), 
\end{align}
% \vspace{-4pt}
\begin{align}
p(\mathcal{A}_t|\mathcal{H}_t,\mathcal{P}_t,\mathcal{T}_t) &= \prod_{j=1}^{|\mathcal{A}_t|} \pi_\theta(\mathcal{A}_t^{j}|\mathcal{H}_t,\mathcal{P}_t,\mathcal{T}_t,\mathcal{A}_t^{<j}). 
\end{align}
% Here, \(\mathcal{T}_t^{j}\) and \(|\mathcal{T}_t|\) denote the \(j\)-th token and the length of \(\mathcal{T}_t\).
% Similarly, \(\mathcal{A}_t^{k}\) and \(|\mathcal{A}_t|\) represent the \(k\)-th token and the length of \(\mathcal{A}_t\), respectively.
% Here, \(\mathcal{T}_t^{j}\) and \(\mathcal{A}_t^{k}\) signify the \(j\)-th and \(k\)-th tokens of \(\mathcal{T}_t\) and \(\mathcal{A}_t\), respectively, with \(|\mathcal{T}_t|\) and \(|\mathcal{A}_t|\) indicating their lengths.
% The model concludes by incorporating the feedback from the action \(\mathcal{A}_t\) as the observation \(\mathcal{O}_t\), which then informs the next historical trajectory \(\mathcal{H}_{t+1}\).
% Lastly, the feedback result of the action \(\mathcal{A}_t\) will be treated as the observation \(\mathcal{O}_t\) and added to the trajectory, generating a new round of historical trajectory \(\mathcal{H}_{t+1}\).

% \end{small}
% This enables the model to effectively apply the knowledge encapsulated in the action knowledge to accomplish inferential tasks.
% This is due to our action rules encompassing the sequential order of actions. 
% Accordingly, the model guides its reasoning path planning based on these established rules, ensuring the generation of a compliant planning path and avoiding irrational path outputs.
% Specifically, in the ActionPath sequence \textit{``Start -> Search -> Lookup -> Finish''}(omitting the details of each operation), the process involves progressing from the initial stage, sequentially selecting ``Search'', then ``Lookup'', before reaching the final ``Finish'' stage.

\begin{algorithm}[t]
\small
\caption{Trajectory Synthesis and Knowledgeable Self-Learning}
\label{algorithm1}
\KwIn{\(AK_m\): Task-specific Action Knowledge from \(Action\;KB\), including a set of actions \(E_a\) and action rules \(R\); \(D_0\): Initial training set; \(D_{\text{test}}\): Testing set; $\epsilon$: Hyperparameter.}

\KwOut{Optimized models \(M = \{M_1, M_2, \ldots\}\).}

\textbf{Initialize:} Model \(M_0\).

\For{\(i = 0\) \textbf{until} test performance stabilizes}{
    \tcpo{Synthesize trajectories.}\par
    \(T_i \leftarrow Traj(M_i,AK_m, D_0)\)\par
    \tcpo{Handle trajectories.}\par
    \If{\(i = 0\)}{
        % \vspace{2pt}
        \tcpo{Fiter trajectories.}\par
        \(T_i' \leftarrow Filter(T_i, AK_m)\)\par
        % \vspace{2pt}
        % \tcpo{Initial fine-tuning.}\par
        % \(M_{i+1} \leftarrow Tune(T_i', M_i)\)
    }
    \Else{
        % \tcpo{Synthesize trajectories.}\par
        % \(T_i \leftarrow Traj(M_i,AK_m, D_0)\)\par
        % \vspace{2pt}
        \tcpo{Filter and merge trajectories.}\par
        \(T_i' \leftarrow FilterAndMerge(T_i,T_{i-1}, AK_m)\)\par
        % \vspace{2pt}
        % \tcpo{Further fine-tuning.}\par
        % \(M_{i+1} \leftarrow Tune(T_i', M_i)\)
    }
    \tcpo{Fine-tuning.}\par
    \(M_{i+1} \leftarrow Tune(T_i', M_i)\)
    \vspace{2pt} \\
    \tcpo{Performance check.}\par
    \vspace{2pt}
    \If{\(\Delta Perf(M_{i+1}, M_i, D_{test}) \leq \epsilon\)}{
        \textbf{break}
    }
}
\Return{Optimized models \(M\)}
\label{algo:self_learning}
\end{algorithm}

% \subsection{Self-Learning with Execution Feedback}
\subsection{Planning Path Refinement via Knowledgeable Self-Learning}
\label{sec:learning_via_path}
In this phase, we introduce \textit{knowledgeable self-learning}.
Our goal is to help the model understand the \textit{action knowledge} more deeply through iterative fine-tuning.
% , enhancing its ability to choose optimal planning paths. 
% This involves introducing a \textit{knowledgeable self-learning} mechanism (Figure~\ref{fig:framework}) that refines the model through data trajectories synthesized in the planning process. 
% By collecting these trajectories for fine-tuning and engaging in iterative training, we enhance the model's performance and efficiency.
As shown in Algorithm \ref{algo:self_learning}, 
our approach begins with an initial training set $D_{0}$ and an untrained model $M_0$, leading to the synthesis of initial trajectories $T_0 = \{\tau_1, \tau_2, \ldots, \tau_n \}$. 
After filtering, these initial outcomes inform further training, producing a preliminary model version, $M_{1}$. 
Subsequently, $M_{1}$ undergoes re-evaluation on $D_{0}$ to create new trajectories $T_1 = \{\tau'_1, \tau'_2, \ldots, \tau'_n \}$.
These trajectories, alongside $T_0$, undergo a filtering and merging process based on \textit{action knowledge}.
This refined set of trajectories is then utilized to fine-tune the model, resulting in an improved version, $M_{2}$.
We continue iterating until the performance improvement on $M_{test}$ becomes small, at which point we halt the iteration process.

\begin{table*}[t!]
\centering
\renewcommand\arraystretch{1.0}
\scalebox{0.85}{
\begin{tabular}{cllcccccc}
\hline
\toprule
{\multirow{2}{*}{\textbf{Backbone}}}
& {\multirow{2}{*}{\textbf{Strategy}}} 
& {\multirow{2}{*}{\textbf{Method}}} 
& \multicolumn{4}{c}{\textbf{HotpotQA}} 
& \multicolumn{2}{c}{\textbf{ALFWorld}}  \\
\cmidrule(lr){4-7} \cmidrule(lr){8-9}
 & & &\textit{Easy} & \textit{Medium} & \textit{Hard} & \textit{Average} & \textit{Seen}& \textit{Unseen} \\
\Xhline{1px}
\multirow{1}{*}{\makecell{GPT-3.5-Turbo}} &Prompting & ReAct~\cite{react} &47.58 &44.38 &34.33 &42.10 &7.86 &  5.97 \\
\multirow{1}{*}{\makecell{GPT-4}} &Prompting & ReAct~\cite{react} &64.39 &66.18 &55.41 &62.00  &43.57 & 38.81  \\
\Xhline{1px}
\multirow{6}{*}{\makecell{Llama-2\\7b-chat}}
     & Prompting
    & CoT~\cite{cot}
     & 35.80 & 26.69 & 18.20 & 26.90 &-  &- \\
     & Prompting
    & ReAct~\cite{react}
     & 25.14 & 19.87 & 17.39 & 20.80 &1.43  & 0.75 \\
     & Prompting
    & Reflexion~\cite{reflexion}
     & 35.55 & 28.73 & 24.35 & 29.54&2.86   &2.24   \\
      & Fine-tuning 
    &  NAT~\cite{nat}
     &39.33  &28.33  &23.33  &30.33 &26.43  &24.52  \\
     & Fine-tuning 
    & FireAct~\cite{fireact}
     & 40.56 & 31.70 & 24.13 & 32.13 & 27.86 & 25.38 \\
     \cmidrule(lr){2-9}
     & Fine-tuning 
     % & Fine-tune
    & \textbf{\ours}-\textcolor{light_blue}{7b}
     &\textbf{40.80} &\textbf{32.49} &\textbf{27.12} &\textbf{33.47} &\textbf{29.26}    &\textbf{29.35}\\
    %  & Fine-tuning 
    % &  {\ours}-\textcolor{light_blue}{7B} (distilled)
    %  &34.30  &20.63  &20.72  &25.22 &18.66 \\
\Xhline{1px}
\multirow{6}{*}{\makecell{Llama-2\\13b-chat}}
    & Prompting
    & CoT~\cite{cot}
    & 37.90 & 25.28 & 21.64 & 28.27 & -  & -   \\
    & Prompting
    & ReAct~\cite{react}
    & 28.68 & 22.15 & 21.69 & 24.17 &13.58  &9.70 \\
    & Prompting
    & Reflexion~\cite{reflexion}
    & 44.43 & 37.50 & 28.17 & 36.70 &18.57  &17.16  \\
   & Fine-tuning 
    &  NAT~\cite{nat}
     & 46.67 &35.33   &28.67  &36.89  &42.38  &46.02  \\
     & Fine-tuning 
    % & Fine-tune 
    & FireAct~\cite{fireact}
    & \textbf{51.95} & 33.93 & 28.88 & 38.26 & 47.86 & 50.37 \\
    \cmidrule(lr){2-9}
     & Fine-tuning 
    % & Fine-tune 
    & \textbf{\ours}-\textcolor{medium_blue}{13b} 
    &46.97 &\textbf{37.60} &\textbf{33.22} &\textbf{39.26}&\textbf{54.29}   &\textbf{58.71} \\
    % &\textbf{52.24}\\
     % & Fine-tuning 
    % & Fine-tune 
    % & {\ours}-\textcolor{medium_blue}{13B} (distilled)
    %  &28.62      &24.70   &23.85   &25.72 &51.49  \\
\Xhline{1.2px}
\multirow{6}{*}{\makecell{Llama-2\\70b-chat}}
    & Prompting
    & CoT ~\cite{cot}
    & 45.37 & 36.33 & 32.27 & 37.99 &- &-\\
    & Prompting
    & ReAct~\cite{react}
    & 39.70 & 37.19 & 33.62 & 36.83 &41.43  &37.31 \\
    & Prompting
    & Reflexion~\cite{reflexion}
    & 48.01 & 46.35 & 35.64 & 43.33 &49.29   &47.01 \\
       & Fine-tuning 
    &  NAT~\cite{nat}
     & 50.66 & 43.67  &44.33   &46.22 &75.95  & 75.62  \\
     & Fine-tuning 
    % & Fine-tune
    & FireAct~\cite{fireact}
    & 51.96 & 47.56 & \textbf{44.60} & 48.04 & \textbf{77.85} & 77.61 \\ 
    \cmidrule(lr){2-9}
     & Fine-tuning 
    % & Fine-tune 
    & \textbf{\ours}-\textcolor{dark_blue}{70b} 
    & \textbf{56.75} & \textbf{49.90} &37.76 &\textbf{48.14} & 77.14   &\textbf{78.36} \\
    %  & Fine-tuning 
    % % & Fine-tune 
    % & {\ours}-\textcolor{dark_blue}{70B} (distilled)
    %  &  &  &  &  \\
\bottomrule
\hline
\end{tabular}}
\caption{\textbf{Overall performance of {\ours} on HotpotQA and ALFWorld.} The evaluation metrics are F1 Score (\%) and Success Rate (\%), respectively. \textbf{Strategy} means the agent learning paradigm behind each method. The best results of each backbone are marked in \textbf{bold}.
}
% \vskip -0.1in
\label{tab:main_results}
\end{table*}

\paragraph{Knowledge-Based Trajectory Filtering and Merging.}
Our \textit{knowledgeable self-learning} approach enhances trajectory quality through two key phases:
% In our \textit{knowledgeable self-learning} process, leveraging action knowledge is crucial for enhancing the quality of trajectories.
% This process involves two key phases:
(1) \textbf{Filtering}:  
We start by selecting correct trajectories, $T_{\text{correct}}$, based on their outcomes. 
Specifically for task HotpotQA, we apply \textit{action knowledge} to further refine these trajectories. 
This refinement involves removing any trajectories that do not align with the provided $AK_m$, particularly those with invalid actions or disordered action sequences.
(2) \textbf{Merging}: 
We then merge trajectories generated by models across different iterations.
For trajectories addressing the same task, 
We refine them based on efficiency, specifically retaining the more efficient (shorter path) trajectories ensuring optimal problem-solving effectiveness.

% In this \textit{knowledgeable self-learning} process, action knowledge is vital. 
% It is employed in the trajectory filtering process, where action knowledge is used to selectively filter trajectories from the set.
% Moreover, during the merging of trajectories, a preference for shorter and more effective paths within $T$ and $T'$ significantly enhances the efficiency of problem-solving strategies. 
% Through this method, the model progressively improves its grasp and application of action knowledge, thereby augmenting the model's efficiency in task execution and increasing its accuracy. 
% In summary, this approach underscores the importance of iterative learning and knowledge-based decision-making in the development of advanced models, fostering a deeper understanding and more effective application of action knowledge.

\section{Experiments}

\subsection{Settings}

We evaluate {\ours} on HotpotQA~\cite{datasets/hotpotqa} and ALFWorld~\cite{datasets/alfworld}. 
We employ \texttt{Llama-2-\{7,13,70\}b-chat}~\citep{llama2-model} as the backbone models, and also apply {\ours} to Vicuna~\citep{fastchat}, Mistral~\citep{mistral}, GPT-3.5-Turbo~\citep{gpt-3.5} and GPT-4~\citep{gpt4}.
We compare {\ours} with various baselines including CoT~\cite{cot},  ReAct~\cite{react},  Reflexion~\cite{reflexion}, FireAct~\cite{fireact} and NAT~\cite{nat}.
% More details about the datasets, evaluation metrics, baselines, and training hyper-parameters of our method can be seen in Appendix \ref{app:exp-set}.
More details about our experiment can be seen in Appendix \ref{app:exp-set}.

\subsection{Main Results}
\begin{figure*}[!t]

\centering %图片居中
\includegraphics[width=1\textwidth]{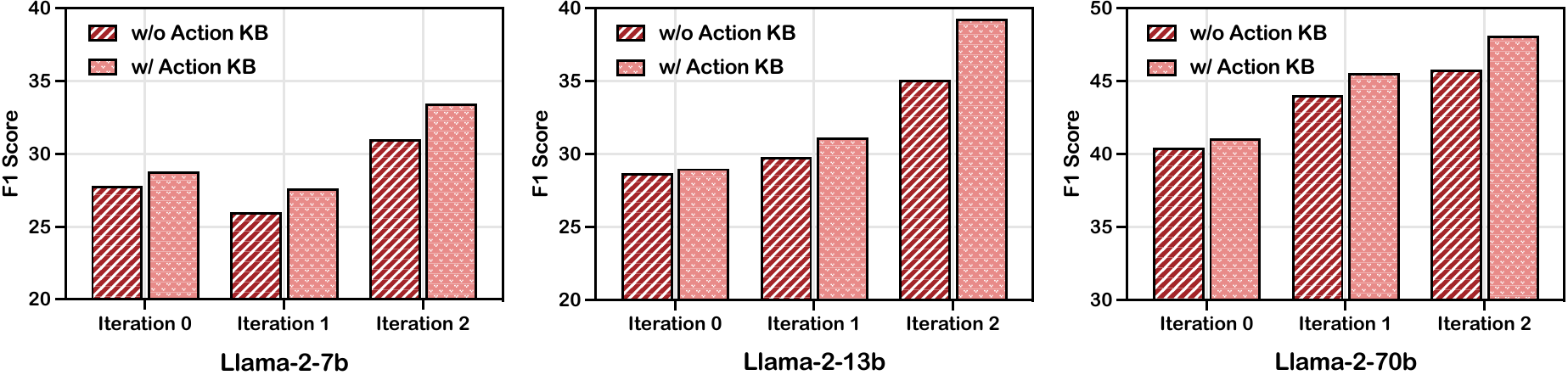} %插入图片，[]中设置图片大小，{}中是图片文件名
\caption{Ablation study on Action Knowledge within Llama-2 Models on HotpotQA. Here \textit{w/ Action KB} indicates the naive {\ours} and \textit{w/o Action KB} symbolizes removing the action knowledge of the specific task.}
% \vskip -0.05in
\label{fig:experiment_ablation}
\end{figure*}

\begin{figure*}[!t]

\centering 
\includegraphics[width=1.\textwidth]{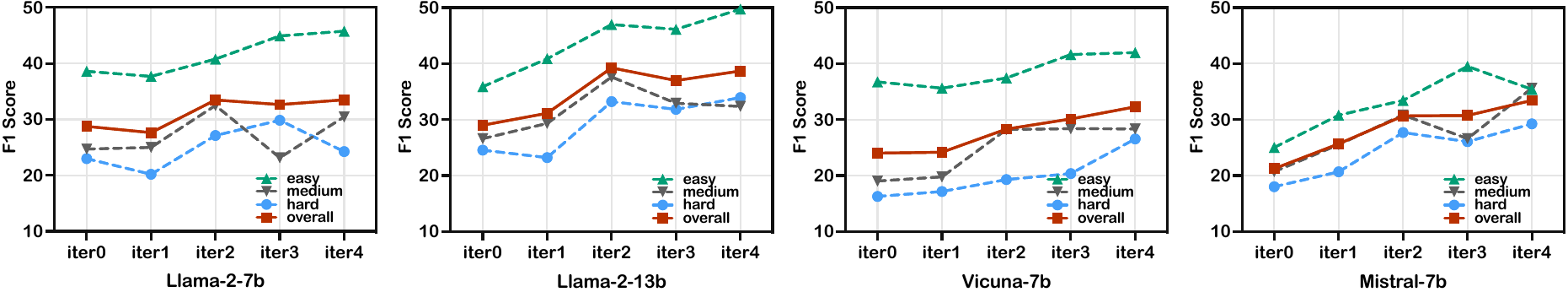} 
\caption{Ablation study on Knowledgeable Self-Learning iteration. We examine the influence of self-learning iterations on a selection of models, including Llama-2-7b, Llama-2-13b, Vicuna-7b, and Mistral-7b. Here \texttt{iter0} represents baseline performance prior to any training. }
\label{fig:experiment_iter}
% \vskip -0.05in
\end{figure*}

\begin{table}[tb]
\centering
\tabcolsep=0.15cm
\scalebox{0.95}{
\begin{tabular}{l|cc}
        \toprule
        {\multirow{2}{*}{\textbf{Method}}} &
        \multicolumn{2}{c}{\textbf{ALFWorld}} \\
                \cmidrule{2-3}
        & \textit{Seen} & \textit{Unseen} \\
                \midrule
ReAct    &13.58   &9.70    \\
 \textbf{\ours}* &30.00 \textcolor{medium_blue}{$_{\uparrow16.42}$}    & 38.06\textcolor{medium_blue}{$_{\uparrow28.36}$}    \\
% \hdashline
\bottomrule
    \end{tabular}}
\caption{Comparative Experiment of ReAct and \ours* on ALFWorld with Llama-2-13b. Here label * denotes the version of \ours ~integrated with action knowledge but not fine-tuned.}
% \vskip -0.15in
\label{tab:ak_analysis}
\end{table}

\paragraph{{\ours} vs. Prompt-based Methods.}
In Table~\ref{tab:main_results}, we present the F1 scores and success rates for {\ours} and various prompt-based methods evaluated on HotpotQA and ALFWorld.
\textbf{Firstly}, across both datasets, {\ours} consistently outperforms the prompt-based baselines on open-source models.
Notably, the 13b model achieves a performance increase of ~\daulg{15.09\%}~(\textit{average}) and ~\daulg{49.01\%}~(\textit{unseen}) over ReAct on the two datasets.
% \todo{}
Additionally, our approach surpasses GPT-4's performance on ALFWorld with both the 13b and 70b models.
% , and outperforms GPT-4 on HotpotQA with the 70b model.
% It also surpasses GPT-4's performance on ALFWorld with both the 13b and 70b models.
% Additionally, our approach surpasses GPT-4’s performance on both HotpotQA and ALFWorld using the 13B model.
\textbf{Secondly}, we conduct additional experiments on ALFWorld, comparing the performance of unrefined \ours*~against ReAct.
The results, shown in Table~\ref{tab:ak_analysis}, validate the effectiveness of \textit{action knowledge} itself. 
Additionally, disparities in effectiveness among different prompt methods are observed, aligning with current research focused on enhancing models’ capabilities to handle complex tasks through diverse strategies such as multi-agent specialization.
\textbf{Specifically}, our investigation is geared towards leveraging external \textit{action knowledge} to facilitate models to more accurately complete complex tasks. 
This is achieved by minimizing invalid actions (on HotpotQA) and promoting action sequences that better reflect real-world situations (on ALFWorld).
% We will offer a detailed discussion later in \S\ref{sec:invalid_action}.
% A more detailed analysis can be found in~\S\ref{sec:invalid_action}.
% Further analysis will be discussed in \S\ref{sec:invalid_action}.
Further analysis, especially in relation to invalid actions in HotpotQA, will be discussed in \S\ref{sec:invalid_action}.

\paragraph{{\ours} vs. Fine-tuning Methods.}
Our comparison here focuses on the fine-tuning results of {\ours} versus FireAct and NAT.
The results also reveal the efficacy of our method. 
% We compare the fine-tuning results of {\ours} with FireAct and NAT, showing that our method is competitive.
Unlike FireAct and NAT, which rely on closed-source models to generate fine-tuning data, {\ours} synthesizes its own.
% Notably, whereas FireAct and NAT rely on strong closed-source models to generate their fine-tuning data, {\ours} synthesizes its own. 
On HotpotQA, for example, FireAct and NAT use 500 trajectories from GPT-4/GPT-3.5-Turbo, while {\ours} selectively fine-tunes on fewer than 300 self-synthesized correct trajectories per iteration. 
This strategy also mirrors in ALFWorld.
The outcomes suggest that self-synthesized data, infused with prior knowledge, can achieve results comparable to data synthesized by more advanced models like GPT-4. 
% Additionally, the results indicate that iterative fine-tuning enables the model to comprehensively grasp the \textit{action knowledge}, leading to superior planning performance.

\begin{figure*}[!t]

\centering 
\includegraphics[width=1.\textwidth]{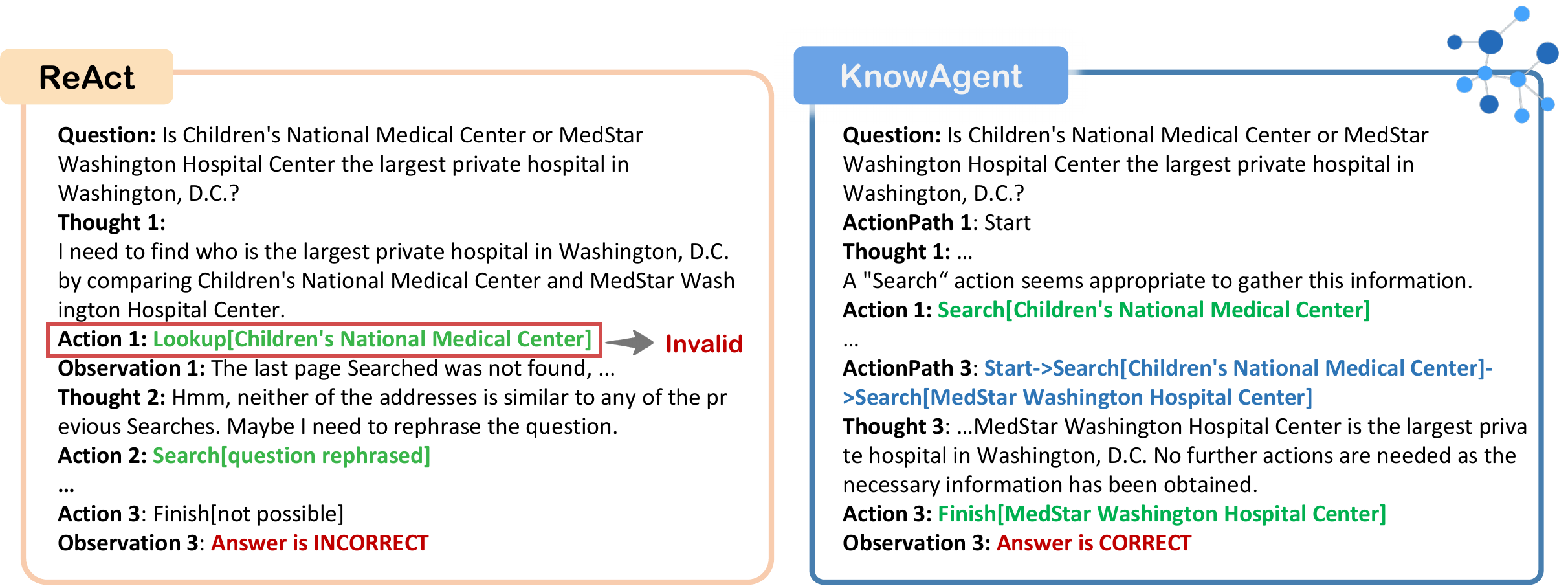} 
\caption{{\ours} \textcolor{medium_blue}{(the blue one)} successfully completes planning, while ReAct \textcolor{lightorange}{(the orange one)} exhibits \textit{planning hallucination} that contradict action knowledge.}
\label{fig:action_impact}
\end{figure*}

\subsection{Analysis}
\label{sec:analysis}
\paragraph{The role of action knowledge grows with the increase of iterations in self-learning.}

Figure~\ref{fig:experiment_ablation} shows the ablation results about \textit{action knowledge} on HotpotQA with Llama series models.
Regardless of the number of iterations, the effect of using \textit{action knowledge} (\textit{w/ action KB}) is superior to that without \textit{action knowledge} (\textit{w/o action KB}).
% , indicating that the introduction of action knowledge can effectively enhance the quality of agent planning.
% The use of action knowledge (\textit{w/ action KB}) yields superior results compared to models without it (\textit{w/o action KB}), regardless of the number of iterations. 
Another interesting finding is that as the number of iterations increases, the performance gap between \textit{w/o action KB} and \textit{w/ action KB} becomes more significant, indicating that the advantages of introducing \textit{action knowledge} become more apparent.
% The use of action knowledge (\textit{w/ action KB}) yields superior results compared to models without it (\textit{w/o action KB}), regardless of the number of iterations. 
% Furthermore, the performance disparity between models with and without action knowledge widens as the number of iterations grows, underscoring the growing advantages of integrating action knowledge. 
We consider that this can be attributed to the virtuous cycle between \textit{action knowledge} and \textit{self-learning}.
Under the constraints of \textit{action knowledge}, the model synthesizes high-quality trajectories for iterative training.
In turn, training on more high-quality trajectories allows the model to better learn \textit{action knowledge}, leading to the generation of even more high-quality trajectories.

\paragraph{Iterative training enhances model proficiency.}
Figure~\ref{fig:experiment_iter} presents a comparative analysis of the effects of iterative training across different models.
\textbf{(1) Number of Iterations.} 
Notably, elevating iterations from one to two results in a substantial optimization of performance.
% Further increasing iterations to 4 leads to an even more pronounced improvement. 
% Extending the iterations further to four continues to yield performance gains, 
Extending the iterations to four continues to enhance results, although the gains become progressively smaller.
These results align with previous studie~\citep{metaai/backtranslation,self-evovled/tuning}, demonstrating the efficacy of iterative self-learning in bolstering the model's comprehension of the training data.
This pattern also reflects the human learning principle of ``Reviewing the old as a means of realizing the new''.
\textbf{(2) Different Base Models.}
We also explore other backbone models except for Llama with a 7b parameter scale, such as Vicuna-7b and Mistral-7b.
The result suggests that our method is effective and generalizable across different pre-trained and fine-tuned models.
Moreover, the performance discrepancies among them also indicate a variance in the ability of different models to absorb and utilize such structured external knowledge.

% \paragraph{Prompting with action knowledge leads to better performance.}

\paragraph{Action knowledge effectively mitigates planning hallucinations.}
\label{sec:invalid_action}
% \begin{table*}[tb]
% \centering
% \tabcolsep=0.15cm
% \scalebox{1}{
% \begin{tabular}{cccc}
% \toprule[1.5pt]
%  Model & Invalid Action &Misordered Action &Duplicate Action\\
% \midrule
% ReAct & 2.08\% &3.54\%    &2.69\%  \\
% Reflexion &  6.87\%&3.87\%  &4.80\% \\
% % FiReAct &  0\%&  0\%&  0\%\\
% KnowAgent & 0.35\%&0\%  &  3.45\% \\
% % \hdashline
% \bottomrule[1.5pt]
% \end{tabular}}
% \caption{Average Invalid Action Proportions for Methods (ReAct, Reflexion, \ours) on Llama-2-7b, Llama-2-13b and Llama-2-70b}
% \label{gpt-mem}
% \end{table*}

\begin{table}[tb]
\centering
\tabcolsep=0.15cm
\scalebox{0.85}{
\begin{tabular}{ccc}
\toprule[1.5pt]
\textbf{Model} & \textbf{Invalid Action} & \textbf{Misordered Action} \\
\midrule
ReAct & 2.08\% & 3.54\%  \\
Reflexion & 6.87\% & 3.87\%  \\
% FiReAct &  0\%&  0\%&  0\%\\
\textbf{\ours} & \textbf{0.35\%} & \textbf{1.23\%}  \\
% \hdashline
\bottomrule[1.5pt]
\end{tabular}}
\caption{Unreasonable action rates on HotpotQA with Llama-2-13b. Here \textit{invalid} refers to actions that do not meet the action rule, while \textit{misordered} means discrepancies in the logical sequence of actions.}
\label{tab:action_rates}
\end{table}
\begin{table}[!t]
\centering
\scalebox{0.85}{
\begin{tabular}{c|c| c c}
\toprule[1.5pt]
\textbf{\ours} & \textbf{Strategy} &\textbf{HotpotQA} & \textbf{ALFWorld}\\
\midrule
  \multirow{2}{*}{\textcolor{light_blue}{7b}}
  &\small{Manual}     &33.47  & 20.15 \\
  & \small{Distilled}   &25.22  & 18.66 \\
 \hline
 \multirow{2}{*}{\textcolor{medium_blue}{13b}}
  & \small{Manual}     &39.26  & 52.24 \\
  &  \small{Distilled}   &25.72  &51.49  \\
\bottomrule[1.5pt]
\end{tabular}
}
\caption{
Comparative Experiment on Manual vs. Distilled Action Knowledge.
\textit{Manual} stands for human-crafted knowledge and \textit{Distilled} represents the distilled knowledge from GPT-4.
}
\label{tab:ablation_manual_distill}
\end{table}

% evaluate action knowledge:
We show the statistical rates of invalid and misordered actions generated by different methods in Table~\ref{tab:action_rates}.
% Here \textit{invalid} refers to actions that do not meet the action rule, while \textit{misordered} means discrepancies in the logical sequence of actions.
% To investigate the impact of incorporating action knowledge into model performance, we conduct a comparative analysis of the ReAct, Reflexion, and {\ours} in Table~\ref{tab:action_analysis}. 
Given that only the \texttt{Search} and \texttt{Finish} actions are involved in FireAct, it is omitted from our analysis here.
% Focusing on the HotpotQA, we perform a statistical analysis of the models' production of invalid actions and misordered actions.
The results 
% in Table~\ref{tab:action_rates} 
demonstrate that incorporating \textit{action knowledge} significantly reduces the frequency of erroneous actions and the likelihood of invalid action paths, thereby increasing the precision of the models on the specific task.
To further substantiate this claim, we refer to the experimental outcomes from {\ours} and ReAct within HotpotQA, as demonstrated in Figure~\ref{fig:action_impact}.
For a given question, ReAct's action sequence follows a \texttt{Lookup->Search} pattern, which is problematic due to the dependency of the \texttt{Lookup} action on the subsequent \texttt{Search} step.
However, with constraints, {\ours} avoids such faulty sequences, enhancing task accuracy.

\paragraph{Distilled vs. Manually Designed Knowledge.}
To investigate whether advanced LLMs can supplant manual efforts in constructing task-specific \textit{action knowledge}, we compare the distilled outcomes from GPT-4 (gpt-4-0613) with our manual-designed ones, using the same actions but different rules~(see Table~\ref{tab:ablation_manual_distill}).
For HotpotQA, the \textit{action knowledge} distilled by GPT-4 is more concise, with fewer cyclical actions than those set by humans.
% While it matches human-defined methods in performance for simpler tasks, it falls short in more complex scenarios requiring longer action sequences. 
This efficiency holds for simpler tasks where performance parallels human-defined methods, while underperformance is found on more complex tasks where longer action sequences are required.
% This efficiency persists in simpler tasks where it achieves similar performance to the manual one, but falls short in complex scenarios requiring longer action sequences.
% Although this distilled knowledge performs comparably to human-defined methods for simpler tasks, it tends to underperform in more complex scenarios requiring longer action sequences. 
For ALFWorld, the GPT-distilled knowledge closely mirrors what is crafted by humans, underscoring the model's capacity to comprehend real-world constraints. 
Aligning with prior research \citep{actionknowledge/robots,self-discover/llm}, this distilled knowledge aids the model in understanding real-world limitations, showing little difference in effectiveness compared to human-created one. 

% \paragraph{Error Analysis.}
% Upon analyzing the capabilities of {\ours}, we identify its limitations, particularly in processing complex queries and summarizing extensive textual data.
% {\ours} struggle to distill key information effectively, often failing to deliver accurate responses.
% The core issue lies in their insufficient reasoning and memory capacities for handling long contexts.
% % which leads to the loss of critical information in text sequences.
% Consequently, the generated responses may be incorrect or even misaligned with the posed questions, such as providing a simple \texttt{yes/no} when a specific entity is required. 
% Future enhancements should focus on strengthening the long-text processing, information retention, and reasoning abilities of our work.

\paragraph{Error Analysis.}
% Upon analyzing the capabilities of {\ours}, we identify its limitations, particularly in processing complex queries and summarizing large texts.
% Due to insufficient reasoning and memory capabilities, {\ours} often generates inaccurate responses when dealing with long contexts, struggling to distill key information effectively.
Upon analyzing the capabilities of {\ours}, we identify its limitations, particularly in processing complex queries and summarizing long texts.
It struggles to distill key information effectively, often failing to respond accurately.
The core issue lies in their insufficient reasoning and memory capacities for handling long contexts.
Consequently, the generated responses may be incorrect or misaligned with the posed questions, such as providing a simple \texttt{yes/no} when a specific entity is required. 
% Our error analysis in HotpotQA, illustrated in Table \ref{tab:error_analysis_cases}, highlights two primary error types: \textbf{Inconsistency Error}, where the model diverges from the question during the answering process, and \textbf{Summarization Error}, where the model fails to extract necessary information from the trajectory, resulting in incorrect answers.
% Future enhancements could focus on improving the model's long-text processing and reasoning abilities.
% Our error analysis in HotpotQA highlights two primary error types: \textbf{Inconsistency Error}, and \textbf{Summarization Error}, and we disscuss it in Appendix~\ref{app: error_analysis}.
% We also offer a detailed comparison of our method with FireAct in terms of model scaling for HotpotQA in Appendix~\ref{app: error_analysis}.
In HotpotQA, we identify two key error types: \textbf{Inconsistency Error} and \textbf{Summarization Error}, discussed further in Appendix~\ref{app: error_analysis}, where we also compare our method to FireAct in terms of model scaling.

\paragraph{Efficiency Analysis.}
Regarding efficiency, taking Llama-2-7b in ALFWorld as an example, {\ours} mandates around 20 hours per cycle, totaling approximately 40 hours for training. 
In comparison, FiReAct, which focuses on training for precise trajectories, requires more training data. 
After incorporating around 710 trajectories synthesized by GPT-4, FireAct's total training time extends to approximately 50 hours.
% Regarding memory usage, both {\ours} and FiReAct have similar memory requirements, around 180GB when running on V100 GPUs.
Regarding memory usage, when running on 8 NVIDIA V100 32G GPUs, both {\ours} and FireAct have similar memory requirements.

\section{Related Work}

\paragraph{LLM-Based Agents.}
LLM-based agents \cite{renda/agent/survey,fudan/agent/survey,multimodel/agent/survey} have emerged as one of the most prevalent AI systems after the rise of LLMs \cite{renda/llm/survey,shuofei/survey,yuqi/survey,personal/agent/survey,kg-agent}.
They learn to interact with the external world through action-observation pairs expressed by natural language.
Previous works primarily focus on unlocking the potential of LLMs as the core of language agents by leveraging human-crafted \cite{react,camel,igas,chatdev} or machine-generated \cite{aiwaves/agents,agentverse,autoagents} prompts.
Recently, there has been a growing emphasis on endowing open-source LLMs with agent capabilities through fine-tuning \cite{lumos,autoact,small/llms/weak/tool}.
However, the training trajectory data for existing language agent fine-tuning methods largely rely on annotations from LLMs.
This can result in the inclusion of trajectories that violate some action knowledge rules and are difficult to identify, leading to an unstable action performance of the trained language agents.
% This can result in the inclusion of trajectories that violate some action knowledge rules, leading to an unstable action performance.
To improve agent performance and reliability, various approaches have been proposed.
Several works focus on designing specialized strategies to enhance agent capabilities.
\citet{amor/multiagent} introduces \textsc{AMOR}, an agent framework that constructs its reasoning capabilities on a Finite State Machine (FSM).
\citet{formal-llm} introduce a ``Formal-LLM'' framework for agents, combining the expressiveness of natural language with the precision of formal language to enhance agent capabilities.
Research efforts~\cite{selfevolve,rerest,actre} have also explored self-improvement approaches, where models continuously enhance their problem-solving abilities through iterative learning in complex tasks.

\paragraph{Knowledge-Augmented LLMs.}
Previous works \cite{ralmpt,ragnlp,atlas} concentrate on knowledge augmentation in LLMs through retrieval.
Due to the rich parameterized knowledge within LLMs~\citep{knowledgemodel/survey,integration/knowledge/llm}, some other works \cite{rainier,generate/rather/retrieve,recitation} advocate for knowledge generation rather than retrieval.
With the emergence of Augmented Language Models (ALMs), many studies \cite{retrieve/cot,cok,freshllms,trice} have enhanced the reasoning capabilities of LLMs by incorporating knowledge from external tools such as search engines, knowledge bases, and Wikipedia documents.
Recent research has explored various approaches to improve LLMs' performance in complex environments: some works \cite{aiwaves/agents,proagent} introduced structured knowledge to regulate multi-agent workflows, while others developed state-aware guidelines for environment-specific reasoning~\cite{stateact,autoguide} or focused on special memory mechanisms for long-horizon tasks~\cite{li2024optimus}.
In this work, we propose knowledge-augmented language agents that incorporate action knowledge rules to constrain the trajectory generation, reducing the occurrence of unreasonable action logic in the generated trajectories.

\section{Conclusion}
In this study, we introduce {\ours}, a framework designed to mitigate \textit{planning hallucinations} by incorporating external action knowledge into synthetic trajectories.
Our method involves utilizing this knowledge to guide the model's action generation and employing a \textit{knowledgeable self-learning} phase for continuous improvement. 
% The approach uses this knowledge to guide action generation, translates it into text for improved model understanding, and incorporates a \textit{knowledgeable self-learning} phase for ongoing enhancement. 
% This multifaceted approach not only enhances the planning capabilities of agents but also proves effective in complex scenarios. 
Our experiments across various models demonstrate that {\ours} effectively competes with or surpasses other baselines, showcasing the benefits of integrating external action knowledge to streamline planning processes and improve performance.

\section*{Limitations}
Our limitations are listed as follows:
\paragraph{Task Expandability.}
The current experiments are conducted exclusively on the commonsense QA and household datasets. However, our approach is also applicable to a broader range of fields including medical~\citep{medagents}, arithmetic~\citep{datasets/gsm8k}, web browsing~\citep{openagents}, and embodied agents~\citep{llm_grounder}. 
This suggests a potential for wider applicability that has yet to be explored.

\paragraph{Multi-Agent Systems.}
Presently, our research focuses on the application of single agents. 
Future studies should explore multi-agent systems, such as \citet{agentverse} and \citet{autoact}, which complete planning tasks through division of labor and collaboration. 
This enhancement could help agents better handle complex tasks and adapt to changing environments.

\paragraph{Automated Design of Action Knowledge Bases.}
The creation of action knowledge bases is still manual, time-consuming, and labor-intensive. 
Even though we use GPT-4 for distilling action knowledge, manual adjustments are needed. 
Future work should aim at automating this process to reduce manual effort and improve the model's autonomous learning and versatility.

% \section*{Ethics Statement}

\section*{Acknowledgments}

We would like to express gratitude to the anonymous reviewers for their kind comments. 
This work was supported by the National Natural Science Foundation of China (No. 62206246, No. NSFCU23B2055, No. NSFCU19B2027), the Fundamental Research Funds for the Central Universities (226-2023-00138), Yongjiang Talent Introduction Programme (2021A-156-G), CIPSC-SMP-Zhipu Large Model Cross-Disciplinary Fund, Ningbo Science and Technology Special Projects under Grant No. 2023Z212, Information Technology Center and State Key Lab of CAD\&CG, Zhejiang University, NUS-NCS Joint Laboratory (A-0008542-00-00), and the Ministry of Education, Singapore, under the Academic Research Fund Tier 1 (FY2023) (Grant A-8001996-00-00).
We gratefully acknowledge the support of Zhejiang University Education Foundation Qizhen Scholar Foundation.
This work was supported by Ant Group and Zhejiang University - Ant Group Joint Laboratory of Knowledge Graph.

\bibliography{custom}

\begin{thebibliography}{81}
\expandafter\ifx\csname natexlab\endcsname\relax\def\natexlab#1{#1}\fi

\bibitem[{Chen et~al.(2023{\natexlab{a}})Chen, Shu, Shareghi, Collier, Narasimhan, and Yao}]{fireact}
Baian Chen, Chang Shu, Ehsan Shareghi, Nigel Collier, Karthik Narasimhan, and Shunyu Yao. 2023{\natexlab{a}}.
\newblock \href {https://doi.org/10.48550/ARXIV.2310.05915} {Fireact: Toward language agent fine-tuning}.
\newblock \emph{CoRR}, abs/2310.05915.

\bibitem[{Chen et~al.(2023{\natexlab{b}})Chen, Dong, Shu, Zhang, Sesay, Karlsson, Fu, and Shi}]{autoagents}
Guangyao Chen, Siwei Dong, Yu~Shu, Ge~Zhang, Jaward Sesay, B{\"{o}}rje~F. Karlsson, Jie Fu, and Yemin Shi. 2023{\natexlab{b}}.
\newblock \href {https://doi.org/10.48550/ARXIV.2309.17288} {Autoagents: {A} framework for automatic agent generation}.
\newblock \emph{CoRR}, abs/2309.17288.

\bibitem[{Chen(2023)}]{knowledgemodel/survey}
Huajun Chen. 2023.
\newblock \href {https://doi.org/10.48550/ARXIV.2312.02706} {Large knowledge model: Perspectives and challenges}.
\newblock \emph{CoRR}, abs/2312.02706.

\bibitem[{Chen et~al.(2023{\natexlab{c}})Chen, Su, Zuo, Yang, Yuan, Qian, Chan, Qin, Lu, Xie, Liu, Sun, and Zhou}]{agentverse}
Weize Chen, Yusheng Su, Jingwei Zuo, Cheng Yang, Chenfei Yuan, Chen Qian, Chi{-}Min Chan, Yujia Qin, Yaxi Lu, Ruobing Xie, Zhiyuan Liu, Maosong Sun, and Jie Zhou. 2023{\natexlab{c}}.
\newblock \href {https://doi.org/10.48550/ARXIV.2308.10848} {Agentverse: Facilitating multi-agent collaboration and exploring emergent behaviors in agents}.
\newblock \emph{CoRR}, abs/2308.10848.

\bibitem[{Cobbe et~al.(2021)Cobbe, Kosaraju, Bavarian, Chen, Jun, Kaiser, Plappert, Tworek, Hilton, Nakano, Hesse, and Schulman}]{datasets/gsm8k}
Karl Cobbe, Vineet Kosaraju, Mohammad Bavarian, Mark Chen, Heewoo Jun, Lukasz Kaiser, Matthias Plappert, Jerry Tworek, Jacob Hilton, Reiichiro Nakano, Christopher Hesse, and John Schulman. 2021.
\newblock \href {http://arxiv.org/abs/2110.14168} {Training verifiers to solve math word problems}.
\newblock \emph{CoRR}, abs/2110.14168.

\bibitem[{Deng et~al.(2023)Deng, Gu, Zheng, Chen, Stevens, Wang, Sun, and Su}]{mind2web}
Xiang Deng, Yu~Gu, Boyuan Zheng, Shijie Chen, Samuel Stevens, Boshi Wang, Huan Sun, and Yu~Su. 2023.
\newblock \href {https://doi.org/10.48550/ARXIV.2306.06070} {Mind2web: Towards a generalist agent for the web}.
\newblock \emph{CoRR}, abs/2306.06070.

\bibitem[{Ding et~al.(2023)Ding, Zhang, Amiri, Cao, Yang, Kaminski, Esselink, and Zhang}]{actionknowledge/robots}
Yan Ding, Xiaohan Zhang, Saeid Amiri, Nieqing Cao, Hao Yang, Andy Kaminski, Chad Esselink, and Shiqi Zhang. 2023.
\newblock \href {https://doi.org/10.1007/S10514-023-10133-5} {Integrating action knowledge and llms for task planning and situation handling in open worlds}.
\newblock \emph{Auton. Robots}, 47(8):981--997.

\bibitem[{Dou et~al.(2024)Dou, Yang, Wu, Chang, and Peng}]{rerest}
Zi{-}Yi Dou, Cheng{-}Fu Yang, Xueqing Wu, Kai{-}Wei Chang, and Nanyun Peng. 2024.
\newblock \href {https://aclanthology.org/2024.emnlp-main.861} {Re-rest: Reflection-reinforced self-training for language agents}.
\newblock In \emph{Proceedings of the 2024 Conference on Empirical Methods in Natural Language Processing, {EMNLP} 2024, Miami, FL, USA, November 12-16, 2024}, pages 15394--15411. Association for Computational Linguistics.

\bibitem[{Durante et~al.(2024)Durante, Huang, Wake, Gong, Park, Sarkar, Taori, Noda, Terzopoulos, Choi, Ikeuchi, Vo, Fei{-}Fei, and Gao}]{multimodel/agent/survey}
Zane Durante, Qiuyuan Huang, Naoki Wake, Ran Gong, Jae~Sung Park, Bidipta Sarkar, Rohan Taori, Yusuke Noda, Demetri Terzopoulos, Yejin Choi, Katsushi Ikeuchi, Hoi Vo, Li~Fei{-}Fei, and Jianfeng Gao. 2024.
\newblock \href {https://doi.org/10.48550/ARXIV.2401.03568} {Agent {AI:} surveying the horizons of multimodal interaction}.
\newblock \emph{CoRR}, abs/2401.03568.

\bibitem[{Feng et~al.(2023)Feng, Ma, Yu, Huang, Wang, Chen, Peng, Feng, Qin, and Liu}]{integration/knowledge/llm}
Zhangyin Feng, Weitao Ma, Weijiang Yu, Lei Huang, Haotian Wang, Qianglong Chen, Weihua Peng, Xiaocheng Feng, Bing Qin, and Ting Liu. 2023.
\newblock \href {https://doi.org/10.48550/ARXIV.2311.05876} {Trends in integration of knowledge and large language models: {A} survey and taxonomy of methods, benchmarks, and applications}.
\newblock \emph{CoRR}, abs/2311.05876.

\bibitem[{Fu et~al.(2024)Fu, Kim, Kim, Sohn, Logeswaran, Bae, and Lee}]{autoguide}
Yao Fu, Dong{-}Ki Kim, Jaekyeom Kim, Sungryull Sohn, Lajanugen Logeswaran, Kyunghoon Bae, and Honglak Lee. 2024.
\newblock Autoguide: Automated generation and selection of state-aware guidelines for large language model agents.
\newblock In \emph{NeurIPS}.

\bibitem[{Guan et~al.(2024)Guan, Wu, Wen, Xu, Wang, and Huang}]{amor/multiagent}
Jian Guan, Wei Wu, Zujie Wen, Peng Xu, Hongning Wang, and Minlie Huang. 2024.
\newblock \href {https://doi.org/10.48550/ARXIV.2402.01469} {{AMOR:} {A} recipe for building adaptable modular knowledge agents through process feedback}.
\newblock \emph{CoRR}, abs/2402.01469.

\bibitem[{Guan et~al.(2023)Guan, Valmeekam, Sreedharan, and Kambhampati}]{llm/world_model/planning}
Lin Guan, Karthik Valmeekam, Sarath Sreedharan, and Subbarao Kambhampati. 2023.
\newblock \href {https://doi.org/10.48550/ARXIV.2305.14909} {Leveraging pre-trained large language models to construct and utilize world models for model-based task planning}.
\newblock \emph{CoRR}, abs/2305.14909.

\bibitem[{Guu et~al.(2020)Guu, Lee, Tung, Pasupat, and Chang}]{ralmpt}
Kelvin Guu, Kenton Lee, Zora Tung, Panupong Pasupat, and Ming{-}Wei Chang. 2020.
\newblock \href {http://proceedings.mlr.press/v119/guu20a.html} {Retrieval augmented language model pre-training}.
\newblock In \emph{Proceedings of the 37th International Conference on Machine Learning, {ICML} 2020, 13-18 July 2020, Virtual Event}, volume 119 of \emph{Proceedings of Machine Learning Research}, pages 3929--3938. {PMLR}.

\bibitem[{Hong et~al.(2023)Hong, Zheng, Chen, Cheng, Wang, Zhang, Wang, Yau, Lin, Zhou, Ran, Xiao, and Wu}]{metagpt}
Sirui Hong, Xiawu Zheng, Jonathan Chen, Yuheng Cheng, Jinlin Wang, Ceyao Zhang, Zili Wang, Steven Ka~Shing Yau, Zijuan Lin, Liyang Zhou, Chenyu Ran, Lingfeng Xiao, and Chenglin Wu. 2023.
\newblock \href {https://doi.org/10.48550/ARXIV.2308.00352} {Metagpt: Meta programming for multi-agent collaborative framework}.
\newblock \emph{CoRR}, abs/2308.00352.

\bibitem[{Ichter et~al.(2022)Ichter, Brohan, Chebotar, Finn, Hausman, Herzog, Ho, Ibarz, Irpan, Jang, Julian, Kalashnikov, Levine, Lu, Parada, Rao, Sermanet, Toshev, Vanhoucke, Xia, Xiao, Xu, Yan, Brown, Ahn, Cortes, Sievers, Tan, Xu, Reyes, Rettinghouse, Quiambao, Pastor, Luu, Lee, Kuang, Jesmonth, Joshi, Jeffrey, Ruano, Hsu, Gopalakrishnan, David, Zeng, and Fu}]{Do_As_I_Can}
Brian Ichter, Anthony Brohan, Yevgen Chebotar, Chelsea Finn, Karol Hausman, Alexander Herzog, Daniel Ho, Julian Ibarz, Alex Irpan, Eric Jang, Ryan Julian, Dmitry Kalashnikov, Sergey Levine, Yao Lu, Carolina Parada, Kanishka Rao, Pierre Sermanet, Alexander Toshev, Vincent Vanhoucke, Fei Xia, Ted Xiao, Peng Xu, Mengyuan Yan, Noah Brown, Michael Ahn, Omar Cortes, Nicolas Sievers, Clayton Tan, Sichun Xu, Diego Reyes, Jarek Rettinghouse, Jornell Quiambao, Peter Pastor, Linda Luu, Kuang{-}Huei Lee, Yuheng Kuang, Sally Jesmonth, Nikhil~J. Joshi, Kyle Jeffrey, Rosario~Jauregui Ruano, Jasmine Hsu, Keerthana Gopalakrishnan, Byron David, Andy Zeng, and Chuyuan~Kelly Fu. 2022.
\newblock \href {https://proceedings.mlr.press/v205/ichter23a.html} {Do as {I} can, not as {I} say: Grounding language in robotic affordances}.
\newblock In \emph{Conference on Robot Learning, CoRL 2022, 14-18 December 2022, Auckland, New Zealand}, volume 205 of \emph{Proceedings of Machine Learning Research}, pages 287--318. {PMLR}.

\bibitem[{Izacard et~al.(2023)Izacard, Lewis, Lomeli, Hosseini, Petroni, Schick, Dwivedi{-}Yu, Joulin, Riedel, and Grave}]{atlas}
Gautier Izacard, Patrick S.~H. Lewis, Maria Lomeli, Lucas Hosseini, Fabio Petroni, Timo Schick, Jane Dwivedi{-}Yu, Armand Joulin, Sebastian Riedel, and Edouard Grave. 2023.
\newblock \href {http://jmlr.org/papers/v24/23-0037.html} {Atlas: Few-shot learning with retrieval augmented language models}.
\newblock \emph{J. Mach. Learn. Res.}, 24:251:1--251:43.

\bibitem[{Jiang et~al.(2023{\natexlab{a}})Jiang, Sablayrolles, Mensch, Bamford, Chaplot, de~Las~Casas, Bressand, Lengyel, Lample, Saulnier, Lavaud, Lachaux, Stock, Scao, Lavril, Wang, Lacroix, and Sayed}]{mistral}
Albert~Q. Jiang, Alexandre Sablayrolles, Arthur Mensch, Chris Bamford, Devendra~Singh Chaplot, Diego de~Las~Casas, Florian Bressand, Gianna Lengyel, Guillaume Lample, Lucile Saulnier, L{\'{e}}lio~Renard Lavaud, Marie{-}Anne Lachaux, Pierre Stock, Teven~Le Scao, Thibaut Lavril, Thomas Wang, Timoth{\'{e}}e Lacroix, and William~El Sayed. 2023{\natexlab{a}}.
\newblock \href {https://doi.org/10.48550/ARXIV.2310.06825} {Mistral 7b}.
\newblock \emph{CoRR}, abs/2310.06825.

\bibitem[{Jiang et~al.(2024)Jiang, Zhou, Zhao, Song, Zhu, Zhu, and Wen}]{kg-agent}
Jinhao Jiang, Kun Zhou, Wayne~Xin Zhao, Yang Song, Chen Zhu, Hengshu Zhu, and Ji{-}Rong Wen. 2024.
\newblock \href {https://doi.org/10.48550/ARXIV.2402.11163} {Kg-agent: An efficient autonomous agent framework for complex reasoning over knowledge graph}.
\newblock \emph{CoRR}, abs/2402.11163.

\bibitem[{Jiang et~al.(2023{\natexlab{b}})Jiang, Wang, and Wang}]{selfevolve}
Shuyang Jiang, Yuhao Wang, and Yu~Wang. 2023{\natexlab{b}}.
\newblock \href {https://doi.org/10.48550/ARXIV.2306.02907} {Selfevolve: {A} code evolution framework via large language models}.
\newblock \emph{CoRR}, abs/2306.02907.

\bibitem[{Lewis et~al.(2020)Lewis, Perez, Piktus, Petroni, Karpukhin, Goyal, K{\"{u}}ttler, Lewis, Yih, Rockt{\"{a}}schel, Riedel, and Kiela}]{ragnlp}
Patrick S.~H. Lewis, Ethan Perez, Aleksandra Piktus, Fabio Petroni, Vladimir Karpukhin, Naman Goyal, Heinrich K{\"{u}}ttler, Mike Lewis, Wen{-}tau Yih, Tim Rockt{\"{a}}schel, Sebastian Riedel, and Douwe Kiela. 2020.
\newblock \href {https://proceedings.neurips.cc/paper/2020/hash/6b493230205f780e1bc26945df7481e5-Abstract.html} {Retrieval-augmented generation for knowledge-intensive {NLP} tasks}.
\newblock In \emph{Advances in Neural Information Processing Systems 33: Annual Conference on Neural Information Processing Systems 2020, NeurIPS 2020, December 6-12, 2020, virtual}.

\bibitem[{Li et~al.(2023{\natexlab{a}})Li, Hammoud, Itani, Khizbullin, and Ghanem}]{camel}
Guohao Li, Hasan Abed Al~Kader Hammoud, Hani Itani, Dmitrii Khizbullin, and Bernard Ghanem. 2023{\natexlab{a}}.
\newblock \href {https://doi.org/10.48550/ARXIV.2303.17760} {{CAMEL:} communicative agents for "mind" exploration of large scale language model society}.
\newblock \emph{CoRR}, abs/2303.17760.

\bibitem[{Li et~al.(2023{\natexlab{b}})Li, Yu, Zhou, Schick, Zettlemoyer, Levy, Weston, and Lewis}]{metaai/backtranslation}
Xian Li, Ping Yu, Chunting Zhou, Timo Schick, Luke Zettlemoyer, Omer Levy, Jason Weston, and Mike Lewis. 2023{\natexlab{b}}.
\newblock \href {https://doi.org/10.48550/ARXIV.2308.06259} {Self-alignment with instruction backtranslation}.
\newblock \emph{CoRR}, abs/2308.06259.

\bibitem[{Li et~al.(2023{\natexlab{c}})Li, Zhao, Chia, Ding, Bing, Joty, and Poria}]{cok}
Xingxuan Li, Ruochen Zhao, Yew~Ken Chia, Bosheng Ding, Lidong Bing, Shafiq~R. Joty, and Soujanya Poria. 2023{\natexlab{c}}.
\newblock \href {https://doi.org/10.48550/ARXIV.2305.13269} {Chain of knowledge: {A} framework for grounding large language models with structured knowledge bases}.
\newblock \emph{CoRR}, abs/2305.13269.

\bibitem[{Li et~al.(2024{\natexlab{a}})Li, Wen, Wang, Li, Yuan, Liu, Liu, Xu, Wang, Sun, Kong, Wang, Geng, Luan, Jin, Ye, Xiong, Zhang, Li, Xu, Li, Li, Liu, Zhang, and Liu}]{personal/agent/survey}
Yuanchun Li, Hao Wen, Weijun Wang, Xiangyu Li, Yizhen Yuan, Guohong Liu, Jiacheng Liu, Wenxing Xu, Xiang Wang, Yi~Sun, Rui Kong, Yile Wang, Hanfei Geng, Jian Luan, Xuefeng Jin, Zilong Ye, Guanjing Xiong, Fan Zhang, Xiang Li, Mengwei Xu, Zhijun Li, Peng Li, Yang Liu, Ya{-}Qin Zhang, and Yunxin Liu. 2024{\natexlab{a}}.
\newblock \href {https://doi.org/10.48550/ARXIV.2401.05459} {Personal {LLM} agents: Insights and survey about the capability, efficiency and security}.
\newblock \emph{CoRR}, abs/2401.05459.

\bibitem[{Li et~al.(2024{\natexlab{b}})Li, Xie, Shao, Chen, Jiang, and Nie}]{li2024optimus}
Zaijing Li, Yuquan Xie, Rui Shao, Gongwei Chen, Dongmei Jiang, and Liqiang Nie. 2024{\natexlab{b}}.
\newblock Optimus-1: Hybrid multimodal memory empowered agents excel in long-horizon tasks.
\newblock In \emph{NeurIPS}.

\bibitem[{Li et~al.(2024{\natexlab{c}})Li, Hua, Wang, Zhu, and Zhang}]{formal-llm}
Zelong Li, Wenyue Hua, Hao Wang, He~Zhu, and Yongfeng Zhang. 2024{\natexlab{c}}.
\newblock \href {https://doi.org/10.48550/ARXIV.2402.00798} {Formal-llm: Integrating formal language and natural language for controllable llm-based agents}.
\newblock \emph{CoRR}, abs/2402.00798.

\bibitem[{Liu et~al.(2022)Liu, Hallinan, Lu, He, Welleck, Hajishirzi, and Choi}]{rainier}
Jiacheng Liu, Skyler Hallinan, Ximing Lu, Pengfei He, Sean Welleck, Hannaneh Hajishirzi, and Yejin Choi. 2022.
\newblock \href {https://doi.org/10.18653/V1/2022.EMNLP-MAIN.611} {Rainier: Reinforced knowledge introspector for commonsense question answering}.
\newblock In \emph{Proceedings of the 2022 Conference on Empirical Methods in Natural Language Processing, {EMNLP} 2022, Abu Dhabi, United Arab Emirates, December 7-11, 2022}, pages 8938--8958. Association for Computational Linguistics.

\bibitem[{Liu et~al.(2023{\natexlab{a}})Liu, Yu, Zhang, Xu, Lei, Lai, Gu, Ding, Men, Yang, Zhang, Deng, Zeng, Du, Zhang, Shen, Zhang, Su, Sun, Huang, Dong, and Tang}]{agentbench}
Xiao Liu, Hao Yu, Hanchen Zhang, Yifan Xu, Xuanyu Lei, Hanyu Lai, Yu~Gu, Hangliang Ding, Kaiwen Men, Kejuan Yang, Shudan Zhang, Xiang Deng, Aohan Zeng, Zhengxiao Du, Chenhui Zhang, Sheng Shen, Tianjun Zhang, Yu~Su, Huan Sun, Minlie Huang, Yuxiao Dong, and Jie Tang. 2023{\natexlab{a}}.
\newblock \href {https://doi.org/10.48550/ARXIV.2308.03688} {Agentbench: Evaluating llms as agents}.
\newblock \emph{CoRR}, abs/2308.03688.

\bibitem[{Liu et~al.(2023{\natexlab{b}})Liu, Yao, Zhang, Xue, Heinecke, Murthy, Feng, Chen, Niebles, Arpit, Xu, Mui, Wang, Xiong, and Savarese}]{bolaa}
Zhiwei Liu, Weiran Yao, Jianguo Zhang, Le~Xue, Shelby Heinecke, Rithesh Murthy, Yihao Feng, Zeyuan Chen, Juan~Carlos Niebles, Devansh Arpit, Ran Xu, Phil Mui, Huan Wang, Caiming Xiong, and Silvio Savarese. 2023{\natexlab{b}}.
\newblock \href {https://doi.org/10.48550/ARXIV.2308.05960} {{BOLAA:} benchmarking and orchestrating llm-augmented autonomous agents}.
\newblock \emph{CoRR}, abs/2308.05960.

\bibitem[{OpenAI(2022)}]{gpt-3.5}
OpenAI. 2022.
\newblock Chatgpt: Optimizing language models for dialogue.
\newblock \url{https://openai.com/blog/chatgpt/}.

\bibitem[{OpenAI(2023)}]{gpt4}
OpenAI. 2023.
\newblock \href {https://doi.org/10.48550/ARXIV.2303.08774} {{GPT-4} technical report}.
\newblock \emph{CoRR}, abs/2303.08774.

\bibitem[{Ouyang and Li(2023)}]{autoplan}
Siqi Ouyang and Lei Li. 2023.
\newblock \href {https://aclanthology.org/2023.findings-emnlp.205} {Autoplan: Automatic planning of interactive decision-making tasks with large language models}.
\newblock In \emph{Findings of the Association for Computational Linguistics: {EMNLP} 2023, Singapore, December 6-10, 2023}, pages 3114--3128. Association for Computational Linguistics.

\bibitem[{Pan et~al.(2023)Pan, Zhai, Yuan, Lv, Fu, Liu, Wang, and Qin}]{kwaiagents}
Haojie Pan, Zepeng Zhai, Hao Yuan, Yaojia Lv, Ruiji Fu, Ming Liu, Zhongyuan Wang, and Bing Qin. 2023.
\newblock \href {https://doi.org/10.48550/ARXIV.2312.04889} {Kwaiagents: Generalized information-seeking agent system with large language models}.
\newblock \emph{CoRR}, abs/2312.04889.

\bibitem[{Qian et~al.(2023)Qian, Cong, Yang, Chen, Su, Xu, Liu, and Sun}]{chatdev}
Chen Qian, Xin Cong, Cheng Yang, Weize Chen, Yusheng Su, Juyuan Xu, Zhiyuan Liu, and Maosong Sun. 2023.
\newblock \href {https://doi.org/10.48550/ARXIV.2307.07924} {Communicative agents for software development}.
\newblock \emph{CoRR}, abs/2307.07924.

\bibitem[{Qiao et~al.(2023{\natexlab{a}})Qiao, Gui, Chen, and Zhang}]{trice}
Shuofei Qiao, Honghao Gui, Huajun Chen, and Ningyu Zhang. 2023{\natexlab{a}}.
\newblock \href {https://doi.org/10.48550/ARXIV.2305.13068} {Making language models better tool learners with execution feedback}.
\newblock \emph{CoRR}, abs/2305.13068.

\bibitem[{Qiao et~al.(2023{\natexlab{b}})Qiao, Ou, Zhang, Chen, Yao, Deng, Tan, Huang, and Chen}]{shuofei/survey}
Shuofei Qiao, Yixin Ou, Ningyu Zhang, Xiang Chen, Yunzhi Yao, Shumin Deng, Chuanqi Tan, Fei Huang, and Huajun Chen. 2023{\natexlab{b}}.
\newblock \href {https://doi.org/10.18653/V1/2023.ACL-LONG.294} {Reasoning with language model prompting: {A} survey}.
\newblock In \emph{Proceedings of the 61st Annual Meeting of the Association for Computational Linguistics (Volume 1: Long Papers), {ACL} 2023, Toronto, Canada, July 9-14, 2023}, pages 5368--5393. Association for Computational Linguistics.

\bibitem[{Qiao et~al.(2024)Qiao, Zhang, Fang, Luo, Zhou, Jiang, Lv, and Chen}]{autoact}
Shuofei Qiao, Ningyu Zhang, Runnan Fang, Yujie Luo, Wangchunshu Zhou, Yuchen~Eleanor Jiang, Chengfei Lv, and Huajun Chen. 2024.
\newblock \href {https://doi.org/10.48550/ARXIV.2401.05268} {{AUTOACT:} automatic agent learning from scratch via self-planning}.
\newblock \emph{CoRR}, abs/2401.05268.

\bibitem[{Rasley et~al.(2020)Rasley, Rajbhandari, Ruwase, and He}]{deepspeed}
Jeff Rasley, Samyam Rajbhandari, Olatunji Ruwase, and Yuxiong He. 2020.
\newblock \href {https://doi.org/10.1145/3394486.3406703} {Deepspeed: System optimizations enable training deep learning models with over 100 billion parameters}.
\newblock In \emph{{KDD} '20: The 26th {ACM} {SIGKDD} Conference on Knowledge Discovery and Data Mining, Virtual Event, CA, USA, August 23-27, 2020}, pages 3505--3506. {ACM}.

\bibitem[{Rozanov and Rei(2024)}]{stateact}
Nikolai Rozanov and Marek Rei. 2024.
\newblock \href {https://doi.org/10.48550/ARXIV.2410.02810} {Stateact: State tracking and reasoning for acting and planning with large language models}.
\newblock \emph{CoRR}, abs/2410.02810.

\bibitem[{Schick et~al.(2023)Schick, Dwivedi{-}Yu, Dess{\`{\i}}, Raileanu, Lomeli, Zettlemoyer, Cancedda, and Scialom}]{toolformer}
Timo Schick, Jane Dwivedi{-}Yu, Roberto Dess{\`{\i}}, Roberta Raileanu, Maria Lomeli, Luke Zettlemoyer, Nicola Cancedda, and Thomas Scialom. 2023.
\newblock \href {https://doi.org/10.48550/ARXIV.2302.04761} {Toolformer: Language models can teach themselves to use tools}.
\newblock \emph{CoRR}, abs/2302.04761.

\bibitem[{Shen et~al.(2024)Shen, Li, Chen, Yan, Quan, Chen, Zhang, and Huang}]{small/llms/weak/tool}
Weizhou Shen, Chenliang Li, Hongzhan Chen, Ming Yan, Xiaojun Quan, Hehong Chen, Ji~Zhang, and Fei Huang. 2024.
\newblock \href {https://doi.org/10.48550/ARXIV.2401.07324} {Small llms are weak tool learners: {A} multi-llm agent}.
\newblock \emph{CoRR}, abs/2401.07324.

\bibitem[{Shinn et~al.(2023)Shinn, Labash, and Gopinath}]{reflexion}
Noah Shinn, Beck Labash, and Ashwin Gopinath. 2023.
\newblock \href {https://doi.org/10.48550/ARXIV.2303.11366} {Reflexion: language agents with verbal reinforcement learning}.
\newblock \emph{CoRR}, abs/2303.11366.

\bibitem[{Shridhar et~al.(2021)Shridhar, Yuan, C{\^{o}}t{\'{e}}, Bisk, Trischler, and Hausknecht}]{datasets/alfworld}
Mohit Shridhar, Xingdi Yuan, Marc{-}Alexandre C{\^{o}}t{\'{e}}, Yonatan Bisk, Adam Trischler, and Matthew~J. Hausknecht. 2021.
\newblock \href {https://openreview.net/forum?id=0IOX0YcCdTn} {Alfworld: Aligning text and embodied environments for interactive learning}.
\newblock In \emph{9th International Conference on Learning Representations, {ICLR} 2021, Virtual Event, Austria, May 3-7, 2021}. OpenReview.net.

\bibitem[{Sumers et~al.(2024)Sumers, Yao, Narasimhan, and Griffiths}]{sumers2024cognitive}
Theodore Sumers, Shunyu Yao, Karthik Narasimhan, and Thomas Griffiths. 2024.
\newblock \href {https://openreview.net/forum?id=1i6ZCvflQJ} {Cognitive architectures for language agents}.
\newblock \emph{Transactions on Machine Learning Research}.
\newblock Survey Certification.

\bibitem[{Sun et~al.(2023{\natexlab{a}})Sun, Zhuang, Kong, Dai, and Zhang}]{adaplanner}
Haotian Sun, Yuchen Zhuang, Lingkai Kong, Bo~Dai, and Chao Zhang. 2023{\natexlab{a}}.
\newblock \href {https://doi.org/10.48550/ARXIV.2305.16653} {Adaplanner: Adaptive planning from feedback with language models}.
\newblock \emph{CoRR}, abs/2305.16653.

\bibitem[{Sun et~al.(2023{\natexlab{b}})Sun, Wang, Tay, Yang, and Zhou}]{recitation}
Zhiqing Sun, Xuezhi Wang, Yi~Tay, Yiming Yang, and Denny Zhou. 2023{\natexlab{b}}.
\newblock \href {https://openreview.net/pdf?id=-cqvvvb-NkI} {Recitation-augmented language models}.
\newblock In \emph{The Eleventh International Conference on Learning Representations, {ICLR} 2023, Kigali, Rwanda, May 1-5, 2023}. OpenReview.net.

\bibitem[{Talebirad and Nadiri(2023)}]{igas}
Yashar Talebirad and Amirhossein Nadiri. 2023.
\newblock \href {https://doi.org/10.48550/ARXIV.2306.03314} {Multi-agent collaboration: Harnessing the power of intelligent {LLM} agents}.
\newblock \emph{CoRR}, abs/2306.03314.

\bibitem[{Tang et~al.(2023)Tang, Zou, Zhang, Zhao, Zhang, Cohan, and Gerstein}]{medagents}
Xiangru Tang, Anni Zou, Zhuosheng Zhang, Yilun Zhao, Xingyao Zhang, Arman Cohan, and Mark Gerstein. 2023.
\newblock \href {https://doi.org/10.48550/ARXIV.2311.10537} {Medagents: Large language models as collaborators for zero-shot medical reasoning}.
\newblock \emph{CoRR}, abs/2311.10537.

\bibitem[{Taori et~al.(2023)Taori, Gulrajani, Zhang, Dubois, Li, Guestrin, Liang, and Hashimoto}]{alpaca-format}
Rohan Taori, Ishaan Gulrajani, Tianyi Zhang, Yann Dubois, Xuechen Li, Carlos Guestrin, Percy Liang, and Tatsunori~B. Hashimoto. 2023.
\newblock Stanford alpaca: An instruction-following llama model.
\newblock \url{https://github.com/tatsu-lab/stanford_alpaca}.

\bibitem[{Team(2023)}]{xagent}
XAgent Team. 2023.
\newblock Xagent: An autonomous agent for complex task solving.

\bibitem[{Touvron et~al.(2023)Touvron, Martin, Stone, Albert, Almahairi, Babaei, Bashlykov, Batra, Bhargava, Bhosale, Bikel, Blecher, Canton{-}Ferrer, Chen, Cucurull, Esiobu, Fernandes, Fu, Fu, Fuller, Gao, Goswami, Goyal, Hartshorn, Hosseini, Hou, Inan, Kardas, Kerkez, Khabsa, Kloumann, Korenev, Koura, Lachaux, Lavril, Lee, Liskovich, Lu, Mao, Martinet, Mihaylov, Mishra, Molybog, Nie, Poulton, Reizenstein, Rungta, Saladi, Schelten, Silva, Smith, Subramanian, Tan, Tang, Taylor, Williams, Kuan, Xu, Yan, Zarov, Zhang, Fan, Kambadur, Narang, Rodriguez, Stojnic, Edunov, and Scialom}]{llama2-model}
Hugo Touvron, Louis Martin, Kevin Stone, Peter Albert, Amjad Almahairi, Yasmine Babaei, Nikolay Bashlykov, Soumya Batra, Prajjwal Bhargava, Shruti Bhosale, Dan Bikel, Lukas Blecher, Cristian Canton{-}Ferrer, Moya Chen, Guillem Cucurull, David Esiobu, Jude Fernandes, Jeremy Fu, Wenyin Fu, Brian Fuller, Cynthia Gao, Vedanuj Goswami, Naman Goyal, Anthony Hartshorn, Saghar Hosseini, Rui Hou, Hakan Inan, Marcin Kardas, Viktor Kerkez, Madian Khabsa, Isabel Kloumann, Artem Korenev, Punit~Singh Koura, Marie{-}Anne Lachaux, Thibaut Lavril, Jenya Lee, Diana Liskovich, Yinghai Lu, Yuning Mao, Xavier Martinet, Todor Mihaylov, Pushkar Mishra, Igor Molybog, Yixin Nie, Andrew Poulton, Jeremy Reizenstein, Rashi Rungta, Kalyan Saladi, Alan Schelten, Ruan Silva, Eric~Michael Smith, Ranjan Subramanian, Xiaoqing~Ellen Tan, Binh Tang, Ross Taylor, Adina Williams, Jian~Xiang Kuan, Puxin Xu, Zheng Yan, Iliyan Zarov, Yuchen Zhang, Angela Fan, Melanie Kambadur, Sharan Narang, Aur{\'{e}}lien Rodriguez, Robert Stojnic, Sergey Edunov,
  and Thomas Scialom. 2023.
\newblock \href {https://doi.org/10.48550/ARXIV.2307.09288} {Llama 2: Open foundation and fine-tuned chat models}.
\newblock \emph{CoRR}, abs/2307.09288.

\bibitem[{Trivedi et~al.(2023)Trivedi, Balasubramanian, Khot, and Sabharwal}]{retrieve/cot}
Harsh Trivedi, Niranjan Balasubramanian, Tushar Khot, and Ashish Sabharwal. 2023.
\newblock \href {https://doi.org/10.18653/V1/2023.ACL-LONG.557} {Interleaving retrieval with chain-of-thought reasoning for knowledge-intensive multi-step questions}.
\newblock In \emph{Proceedings of the 61st Annual Meeting of the Association for Computational Linguistics (Volume 1: Long Papers), {ACL} 2023, Toronto, Canada, July 9-14, 2023}, pages 10014--10037. Association for Computational Linguistics.

\bibitem[{Valmeekam et~al.(2023)Valmeekam, Marquez, and Kambhampati}]{llm/self-critic/plans}
Karthik Valmeekam, Matthew Marquez, and Subbarao Kambhampati. 2023.
\newblock \href {https://doi.org/10.48550/ARXIV.2310.08118} {Can large language models really improve by self-critiquing their own plans?}
\newblock \emph{CoRR}, abs/2310.08118.

\bibitem[{Vu et~al.(2023)Vu, Iyyer, Wang, Constant, Wei, Wei, Tar, Sung, Zhou, Le, and Luong}]{freshllms}
Tu~Vu, Mohit Iyyer, Xuezhi Wang, Noah Constant, Jerry~W. Wei, Jason Wei, Chris Tar, Yun{-}Hsuan Sung, Denny Zhou, Quoc~V. Le, and Thang Luong. 2023.
\newblock \href {https://doi.org/10.48550/ARXIV.2310.03214} {Freshllms: Refreshing large language models with search engine augmentation}.
\newblock \emph{CoRR}, abs/2310.03214.

\bibitem[{Wang et~al.(2023{\natexlab{a}})Wang, Xie, Jiang, Mandlekar, Xiao, Zhu, Fan, and Anandkumar}]{voyager}
Guanzhi Wang, Yuqi Xie, Yunfan Jiang, Ajay Mandlekar, Chaowei Xiao, Yuke Zhu, Linxi Fan, and Anima Anandkumar. 2023{\natexlab{a}}.
\newblock \href {https://doi.org/10.48550/ARXIV.2305.16291} {Voyager: An open-ended embodied agent with large language models}.
\newblock \emph{CoRR}, abs/2305.16291.

\bibitem[{Wang et~al.(2023{\natexlab{b}})Wang, Ma, Feng, Zhang, Yang, Zhang, Chen, Tang, Chen, Lin, Zhao, Wei, and Wen}]{renda/agent/survey}
Lei Wang, Chen Ma, Xueyang Feng, Zeyu Zhang, Hao Yang, Jingsen Zhang, Zhiyuan Chen, Jiakai Tang, Xu~Chen, Yankai Lin, Wayne~Xin Zhao, Zhewei Wei, and Ji{-}Rong Wen. 2023{\natexlab{b}}.
\newblock \href {https://doi.org/10.48550/ARXIV.2308.11432} {A survey on large language model based autonomous agents}.
\newblock \emph{CoRR}, abs/2308.11432.

\bibitem[{Wang et~al.(2024)Wang, Li, Han, Zhang, and Baldwin}]{nat}
Renxi Wang, Haonan Li, Xudong Han, Yixuan Zhang, and Timothy Baldwin. 2024.
\newblock \href {https://doi.org/10.48550/ARXIV.2402.11651} {Learning from failure: Integrating negative examples when fine-tuning large language models as agents}.
\newblock \emph{CoRR}, abs/2402.11651.

\bibitem[{Wei et~al.(2022)Wei, Wang, Schuurmans, Bosma, Ichter, Xia, Chi, Le, and Zhou}]{cot}
Jason Wei, Xuezhi Wang, Dale Schuurmans, Maarten Bosma, Brian Ichter, Fei Xia, Ed~H. Chi, Quoc~V. Le, and Denny Zhou. 2022.
\newblock \href {http://papers.nips.cc/paper\_files/paper/2022/hash/9d5609613524ecf4f15af0f7b31abca4-Abstract-Conference.html} {Chain-of-thought prompting elicits reasoning in large language models}.
\newblock In \emph{NeurIPS}.

\bibitem[{Wu et~al.(2023)Wu, Lu, Xu, Lin, Su, and Zhou}]{self-evovled/tuning}
Shengguang Wu, Keming Lu, Benfeng Xu, Junyang Lin, Qi~Su, and Chang Zhou. 2023.
\newblock \href {https://doi.org/10.48550/ARXIV.2311.08182} {Self-evolved diverse data sampling for efficient instruction tuning}.
\newblock \emph{CoRR}, abs/2311.08182.

\bibitem[{Xi et~al.(2023)Xi, Chen, Guo, He, Ding, Hong, Zhang, Wang, Jin, Zhou, Zheng, Fan, Wang, Xiong, Zhou, Wang, Jiang, Zou, Liu, Yin, Dou, Weng, Cheng, Zhang, Qin, Zheng, Qiu, Huan, and Gui}]{fudan/agent/survey}
Zhiheng Xi, Wenxiang Chen, Xin Guo, Wei He, Yiwen Ding, Boyang Hong, Ming Zhang, Junzhe Wang, Senjie Jin, Enyu Zhou, Rui Zheng, Xiaoran Fan, Xiao Wang, Limao Xiong, Yuhao Zhou, Weiran Wang, Changhao Jiang, Yicheng Zou, Xiangyang Liu, Zhangyue Yin, Shihan Dou, Rongxiang Weng, Wensen Cheng, Qi~Zhang, Wenjuan Qin, Yongyan Zheng, Xipeng Qiu, Xuanjing Huan, and Tao Gui. 2023.
\newblock \href {https://doi.org/10.48550/ARXIV.2309.07864} {The rise and potential of large language model based agents: {A} survey}.
\newblock \emph{CoRR}, abs/2309.07864.

\bibitem[{Xie et~al.(2023)Xie, Zhou, Cheng, Shi, Weng, Liu, Hua, Zhao, Liu, Liu, Liu, Xu, Su, Shin, Xiong, and Yu}]{openagents}
Tianbao Xie, Fan Zhou, Zhoujun Cheng, Peng Shi, Luoxuan Weng, Yitao Liu, Toh~Jing Hua, Junning Zhao, Qian Liu, Che Liu, Leo~Z. Liu, Yiheng Xu, Hongjin Su, Dongchan Shin, Caiming Xiong, and Tao Yu. 2023.
\newblock \href {https://doi.org/10.48550/ARXIV.2310.10634} {Openagents: An open platform for language agents in the wild}.
\newblock \emph{CoRR}, abs/2310.10634.

\bibitem[{Yang et~al.(2023)Yang, Chen, Qian, Madaan, Iyengar, Fouhey, and Chai}]{llm_grounder}
Jianing Yang, Xuweiyi Chen, Shengyi Qian, Nikhil Madaan, Madhavan Iyengar, David~F. Fouhey, and Joyce Chai. 2023.
\newblock \href {https://doi.org/10.48550/ARXIV.2309.12311} {Llm-grounder: Open-vocabulary 3d visual grounding with large language model as an agent}.
\newblock \emph{CoRR}, abs/2309.12311.

\bibitem[{Yang et~al.(2024{\natexlab{a}})Yang, Liu, Wu, Yang, Fung, Li, Huang, Cao, Wang, Wang, Ji, and Zhai}]{code/agent/survey}
Ke~Yang, Jiateng Liu, John Wu, Chaoqi Yang, Yi~R. Fung, Sha Li, Zixuan Huang, Xu~Cao, Xingyao Wang, Yiquan Wang, Heng Ji, and Chengxiang Zhai. 2024{\natexlab{a}}.
\newblock \href {https://doi.org/10.48550/ARXIV.2401.00812} {If {LLM} is the wizard, then code is the wand: {A} survey on how code empowers large language models to serve as intelligent agents}.
\newblock \emph{CoRR}, abs/2401.00812.

\bibitem[{Yang et~al.(2018)Yang, Qi, Zhang, Bengio, Cohen, Salakhutdinov, and Manning}]{datasets/hotpotqa}
Zhilin Yang, Peng Qi, Saizheng Zhang, Yoshua Bengio, William~W. Cohen, Ruslan Salakhutdinov, and Christopher~D. Manning. 2018.
\newblock \href {https://doi.org/10.18653/V1/D18-1259} {Hotpotqa: {A} dataset for diverse, explainable multi-hop question answering}.
\newblock In \emph{Proceedings of the 2018 Conference on Empirical Methods in Natural Language Processing, Brussels, Belgium, October 31 - November 4, 2018}, pages 2369--2380. Association for Computational Linguistics.

\bibitem[{Yang et~al.(2024{\natexlab{b}})Yang, Li, Yan, Zhang, Huang, and Liu}]{actre}
Zonghan Yang, Peng Li, Ming Yan, Ji~Zhang, Fei Huang, and Yang Liu. 2024{\natexlab{b}}.
\newblock \href {https://doi.org/10.48550/ARXIV.2403.14589} {React meets actre: When language agents enjoy training data autonomy}.
\newblock \emph{CoRR}, abs/2403.14589.

\bibitem[{Yao et~al.(2022)Yao, Chen, Yang, and Narasimhan}]{webshop}
Shunyu Yao, Howard Chen, John Yang, and Karthik Narasimhan. 2022.
\newblock \href {http://papers.nips.cc/paper\_files/paper/2022/hash/82ad13ec01f9fe44c01cb91814fd7b8c-Abstract-Conference.html} {Webshop: Towards scalable real-world web interaction with grounded language agents}.
\newblock In \emph{Advances in Neural Information Processing Systems 35: Annual Conference on Neural Information Processing Systems 2022, NeurIPS 2022, New Orleans, LA, USA, November 28 - December 9, 2022}.

\bibitem[{Yao et~al.(2023{\natexlab{a}})Yao, Yu, Zhao, Shafran, Griffiths, Cao, and Narasimhan}]{tot}
Shunyu Yao, Dian Yu, Jeffrey Zhao, Izhak Shafran, Thomas~L. Griffiths, Yuan Cao, and Karthik Narasimhan. 2023{\natexlab{a}}.
\newblock \href {https://doi.org/10.48550/ARXIV.2305.10601} {Tree of thoughts: Deliberate problem solving with large language models}.
\newblock \emph{CoRR}, abs/2305.10601.

\bibitem[{Yao et~al.(2023{\natexlab{b}})Yao, Zhao, Yu, Du, Shafran, Narasimhan, and Cao}]{react}
Shunyu Yao, Jeffrey Zhao, Dian Yu, Nan Du, Izhak Shafran, Karthik~R. Narasimhan, and Yuan Cao. 2023{\natexlab{b}}.
\newblock \href {https://openreview.net/pdf?id=WE\_vluYUL-X} {React: Synergizing reasoning and acting in language models}.
\newblock In \emph{The Eleventh International Conference on Learning Representations, {ICLR} 2023, Kigali, Rwanda, May 1-5, 2023}. OpenReview.net.

\bibitem[{Yao et~al.(2023{\natexlab{c}})Yao, Heinecke, Niebles, Liu, Feng, Xue, Murthy, Chen, Zhang, Arpit, Xu, Mui, Wang, Xiong, and Savarese}]{retroformer}
Weiran Yao, Shelby Heinecke, Juan~Carlos Niebles, Zhiwei Liu, Yihao Feng, Le~Xue, Rithesh Murthy, Zeyuan Chen, Jianguo Zhang, Devansh Arpit, Ran Xu, Phil Mui, Huan Wang, Caiming Xiong, and Silvio Savarese. 2023{\natexlab{c}}.
\newblock \href {https://doi.org/10.48550/ARXIV.2308.02151} {Retroformer: Retrospective large language agents with policy gradient optimization}.
\newblock \emph{CoRR}, abs/2308.02151.

\bibitem[{Ye et~al.(2023)Ye, Cong, Tian, Cao, Wang, Qin, Lu, Yu, Wang, Lin, Liu, and Sun}]{proagent}
Yining Ye, Xin Cong, Shizuo Tian, Jiannan Cao, Hao Wang, Yujia Qin, Yaxi Lu, Heyang Yu, Huadong Wang, Yankai Lin, Zhiyuan Liu, and Maosong Sun. 2023.
\newblock \href {https://doi.org/10.48550/ARXIV.2311.10751} {Proagent: From robotic process automation to agentic process automation}.
\newblock \emph{CoRR}, abs/2311.10751.

\bibitem[{Yin et~al.(2023)Yin, Brahman, Ravichander, Chandu, Chang, Choi, and Lin}]{lumos}
Da~Yin, Faeze Brahman, Abhilasha Ravichander, Khyathi Chandu, Kai{-}Wei Chang, Yejin Choi, and Bill~Yuchen Lin. 2023.
\newblock \href {https://doi.org/10.48550/ARXIV.2311.05657} {Lumos: Learning agents with unified data, modular design, and open-source llms}.
\newblock \emph{CoRR}, abs/2311.05657.

\bibitem[{Yu et~al.(2023)Yu, Iter, Wang, Xu, Ju, Sanyal, Zhu, Zeng, and Jiang}]{generate/rather/retrieve}
Wenhao Yu, Dan Iter, Shuohang Wang, Yichong Xu, Mingxuan Ju, Soumya Sanyal, Chenguang Zhu, Michael Zeng, and Meng Jiang. 2023.
\newblock \href {https://openreview.net/pdf?id=fB0hRu9GZUS} {Generate rather than retrieve: Large language models are strong context generators}.
\newblock In \emph{The Eleventh International Conference on Learning Representations, {ICLR} 2023, Kigali, Rwanda, May 1-5, 2023}. OpenReview.net.

\bibitem[{Zeng et~al.(2023)Zeng, Liu, Lu, Wang, Liu, Dong, and Tang}]{agenttuning}
Aohan Zeng, Mingdao Liu, Rui Lu, Bowen Wang, Xiao Liu, Yuxiao Dong, and Jie Tang. 2023.
\newblock \href {https://doi.org/10.48550/ARXIV.2310.12823} {Agenttuning: Enabling generalized agent abilities for llms}.
\newblock \emph{CoRR}, abs/2310.12823.

\bibitem[{Zhang et~al.(2023)Zhang, Yao, Zhang, Tang, Ma, He, Wang, Gerstein, Wang, Liu, and Zhao}]{sjtu/cot/survey}
Zhuosheng Zhang, Yao Yao, Aston Zhang, Xiangru Tang, Xinbei Ma, Zhiwei He, Yiming Wang, Mark Gerstein, Rui Wang, Gongshen Liu, and Hai Zhao. 2023.
\newblock \href {https://doi.org/10.48550/ARXIV.2311.11797} {Igniting language intelligence: The hitchhiker's guide from chain-of-thought reasoning to language agents}.
\newblock \emph{CoRR}, abs/2311.11797.

\bibitem[{Zhao et~al.(2023)Zhao, Zhou, Li, Tang, Wang, Hou, Min, Zhang, Zhang, Dong, Du, Yang, Chen, Chen, Jiang, Ren, Li, Tang, Liu, Liu, Nie, and Wen}]{renda/llm/survey}
Wayne~Xin Zhao, Kun Zhou, Junyi Li, Tianyi Tang, Xiaolei Wang, Yupeng Hou, Yingqian Min, Beichen Zhang, Junjie Zhang, Zican Dong, Yifan Du, Chen Yang, Yushuo Chen, Zhipeng Chen, Jinhao Jiang, Ruiyang Ren, Yifan Li, Xinyu Tang, Zikang Liu, Peiyu Liu, Jian{-}Yun Nie, and Ji{-}Rong Wen. 2023.
\newblock \href {https://doi.org/10.48550/ARXIV.2303.18223} {A survey of large language models}.
\newblock \emph{CoRR}, abs/2303.18223.

\bibitem[{Zheng et~al.(2023)Zheng, Chiang, Sheng, Zhuang, Wu, Zhuang, Lin, Li, Li, Xing, Zhang, Gonzalez, and Stoica}]{fastchat}
Lianmin Zheng, Wei-Lin Chiang, Ying Sheng, Siyuan Zhuang, Zhanghao Wu, Yonghao Zhuang, Zi~Lin, Zhuohan Li, Dacheng Li, Eric.~P Xing, Hao Zhang, Joseph~E. Gonzalez, and Ion Stoica. 2023.
\newblock \href {http://arxiv.org/abs/2306.05685} {Judging llm-as-a-judge with mt-bench and chatbot arena}.

\bibitem[{Zhou et~al.(2024)Zhou, Pujara, Ren, Chen, Cheng, Le, Chi, Zhou, Mishra, and Zheng}]{self-discover/llm}
Pei Zhou, Jay Pujara, Xiang Ren, Xinyun Chen, Heng{-}Tze Cheng, Quoc~V. Le, Ed~H. Chi, Denny Zhou, Swaroop Mishra, and Huaixiu~Steven Zheng. 2024.
\newblock \href {https://doi.org/10.48550/ARXIV.2402.03620} {Self-discover: Large language models self-compose reasoning structures}.
\newblock \emph{CoRR}, abs/2402.03620.

\bibitem[{Zhou et~al.(2023{\natexlab{a}})Zhou, Xu, Zhu, Zhou, Lo, Sridhar, Cheng, Bisk, Fried, Alon, and Neubig}]{webarena}
Shuyan Zhou, Frank~F. Xu, Hao Zhu, Xuhui Zhou, Robert Lo, Abishek Sridhar, Xianyi Cheng, Yonatan Bisk, Daniel Fried, Uri Alon, and Graham Neubig. 2023{\natexlab{a}}.
\newblock \href {https://doi.org/10.48550/ARXIV.2307.13854} {Webarena: {A} realistic web environment for building autonomous agents}.
\newblock \emph{CoRR}, abs/2307.13854.

\bibitem[{Zhou et~al.(2023{\natexlab{b}})Zhou, Jiang, Li, Wu, Wang, Qiu, Zhang, Chen, Wu, Wang, Zhu, Chen, Zhang, Zhang, Chen, Cui, and Sachan}]{aiwaves/agents}
Wangchunshu Zhou, Yuchen~Eleanor Jiang, Long Li, Jialong Wu, Tiannan Wang, Shi Qiu, Jintian Zhang, Jing Chen, Ruipu Wu, Shuai Wang, Shiding Zhu, Jiyu Chen, Wentao Zhang, Ningyu Zhang, Huajun Chen, Peng Cui, and Mrinmaya Sachan. 2023{\natexlab{b}}.
\newblock \href {https://doi.org/10.48550/ARXIV.2309.07870} {Agents: An open-source framework for autonomous language agents}.
\newblock \emph{CoRR}, abs/2309.07870.

\bibitem[{Zhu et~al.(2023)Zhu, Wang, Chen, Qiao, Ou, Yao, Deng, Chen, and Zhang}]{yuqi/survey}
Yuqi Zhu, Xiaohan Wang, Jing Chen, Shuofei Qiao, Yixin Ou, Yunzhi Yao, Shumin Deng, Huajun Chen, and Ningyu Zhang. 2023.
\newblock \href {https://doi.org/10.48550/ARXIV.2305.13168} {Llms for knowledge graph construction and reasoning: Recent capabilities and future opportunities}.
\newblock \emph{CoRR}, abs/2305.13168.

\end{thebibliography}
\bibliographystyle{acl_natbib}

\appendix

\section{Experimental Settings}
\label{app:exp-set}
\paragraph{Datasets and Metrics.}
HotpotQA~\cite{datasets/hotpotqa} is specifically designed for multi-hop reasoning tasks and comprises approximately 113,000 question-answering pairs derived from Wikipedia articles. 
In our experiments, a subset of 500 data instances is randomly sampled for training.
For testing, we utilize the same test set as BOLAA~\cite{bolaa}, which included 100 samples at each of three difficulty levels: easy, medium and hard.
To measure performance, we adopt the F1 score as a benchmark, which also serves as the basis for the reward metric utilized in BOLAA.
More details of HotpotQA are listed in Appendix~\ref{app: datasets}.

ALFWorld~\cite{datasets/alfworld} is an interactive, text-based household environment where agent challenge to complete six different types of multi-step tasks. 
To train our model, we randomly selected 85 instances from each task category in the training set. 
Furthermore, following previous research~\citep{datasets/alfworld,react}, we use a test set with 140 seen tasks and 134 unseen tasks to evaluate our method.
In ALFWorld, we utilize goal-conditioned success rates as our evaluation metric.
We show prompts used for both datasets in Appendix~\ref{app:prompt}.

\paragraph{Baselines}
Our research concentrates solely on the performance of a single agent, intentionally excluding multi-agent studies from our baseline comparison. 
We choose the following baselines here:
(1) Chain-of-Thought (CoT)~\cite{cot} catalyzes in-depth reasoning in large language models (LLMs) by incorporating intermediate reasoning steps within examples.
% , thereby enhancing result accuracy. 
(2) ReAct~\cite{react} enables LLMs to intertwine the generation of inferential trajectories and actions, allowing the model to better generalize and adjust its action plans.
% Moreover, it facilitates interaction with the environment through these actions to gather additional information, thereby achieving complex reasoning steps.
(3) The Reflexion~\cite{reflexion} method utilizes self-reflective feedback to promote continuous agent development through the assimilation of lessons from past errors, thus refining task execution.
% , by endowing agents with dynamic memory and self-reflection capabilities, fortifies their ability to act along reasoning traces and make task-relevant choices. 
% Feedback from self-reflection guides the agents towards improvement, allowing them to learn from past errors and better accomplish tasks.
(4) FireAct~\cite{fireact} introduces a novel strategy to fine-tune LLM using diverse prompts and trajectories across tasks, demonstrating that richer fine-tuning data can further elevate the performance of agents.
(5) Furthermore, NAT~\cite{nat} is a method to enhance agent performance by leveraging both successful and failed trajectories during fine-tuning, showing that learning from failures can significantly boost model effectiveness.
% Our experiments focus on the effect of external action knowledge on LLMs, therefore we only select single-agent frameworks here as the baseline.
It's important to note that our experiments concentrate on exploring the impact of external action knowledge on LLMs, and as such, we have chosen a single-agent framework as the baseline for comparison.
\paragraph{Implementation.}
We evaluate {\ours} on HotpotQA~\cite{datasets/hotpotqa} and ALFWorld~\cite{datasets/alfworld}. 
We employ \texttt{Llama-2-\{7,13,70\}b-chat}~\citep{llama2-model} as the backbone models,
% and also apply {\ours} to 
% vicuna-7b-v1.5-16k~\citep{fastchat} and Mistral-7B-Instruct-v0.1~\citep{mistral}.
and also apply {\ours} to 
vicuna-7b-v1.5-16k~\citep{fastchat}, Mistral-7B-Instruct-v0.1~\citep{mistral}, gpt-3.5-turbo-0125~\cite{gpt-3.5} and gpt-4-32K-0613.
During the \textit{self-learning} phase, we set the number of training iterations to two. 
% We fine-tune these models over 5 epochs using LoRA~\cite{train/lora}, setting a learning rate of $1 \times 10^{-4}$ and the sequence length of 4096 across all scales. 
Additionally, we employ an Alpaca-style~\cite{alpaca-format} templating approach for generating input from instructional data.
Our fine-tuning framework leverages FastChat \cite{fastchat} using DeepSpeed \cite{deepspeed}.
% And all of our experiments are conducted on V100 GPUs.
Most of our experiments are conducted using NVIDIA V100 32G GPUs. However, for the experiments involving the 13b and 70b models in the ALFWorld environment, we utilize NVIDIA A800 80G GPUs.
We detail the hyper-parameters for training in Table~\ref{tab:hp}, with a default batch size of 2.

\begin{table*}[t!]
    \centering
    \renewcommand\arraystretch{1}
    \scalebox{1.}{
    \begin{tabular}{ccc}
    \toprule
    \textbf{Name} &  \textbf{Value} \\
    \Xhline{1px}
    lora\_r & 8  \\
    lora\_alpha & 16 \\
    lora\_dropout & 0.05 \\
    lora\_target\_modules & q\_proj, v\_proj \\
    model\_max\_length & 4096 \\
    per\_device\_batch\_size & 2  \\
    gradient\_accumulation\_steps & 1 \\
    warmup\_ratio & 0.03 \\
    epochs & 5 \\
    % batch size & 4\\
    learning rate & 1e-4\\
    \bottomrule
    \end{tabular}
    }
    \caption{Hyper-parameters used in our paper.}
    \label{tab:hp}
\end{table*}

\section{Datasets}
\label{app: datasets}
In HotpotQA, we have enriched the action set in previous works~\citep{react,bolaa} by incorporating Bing Search as an external knowledge source. 
As a result, we now feature four distinct actions within HotpotQA to enhance its functionality and performance:
(1) Retrieve[entity]: Retrieve the exact entity on Wikipedia and return the first paragraph if it exists. If not, return some similar entities for searching. 
(2) Search[topic]: Use Bing Search to find relevant information on a specified topic, question, or term.
(3) Lookup[keyword]: Return the next sentence that contains the keyword in the last passage successfully found by Search or Retrieve.  
(4) Finish[answer]: Return the answer and conclude the task.

% \section{Impact of Action Knowledge}
% \label{app:impact_of_ak}
% See Figure~\ref{fig:action_impact}.
% \input{images/error_case}

\section{Error Analysis}
\label{app: error_analysis}
Our error analysis in HotpotQA, illustrated in Table \ref{tab:error_analysis_cases}, highlights two primary error types: \textbf{Inconsistency Error}, where the model diverges from the question during the answering process, and \textbf{Summarization Error}, where the model fails to extract necessary information from the trajectory, resulting in incorrect answers.
Future enhancements could focus on improving the model's long-text processing and reasoning abilities.
% Our error analysis in HotpotQA highlights two primary error types: \textbf{Inconsistency Error}, and \textbf{Summarization Error}, and we disscuss it in Appendix~\ref{app: error_analysis}.
% We also offer a detailed comparison of our method with FireAct in terms of model scaling for HotpotQA in Appendix~\ref{app: error_analysis}.

We also observe different performance trends between our method and FireAct when scaling the model size for HotpotQA.
For the 7b model, {\ours} performs better than FireAct because it effectively avoids plan hallucination, which significantly improves task outcomes.
With the 13b model, we observe that FireAct performs better on easy tasks. 
This suggests that using more and higher-quality trajectories (synthesized by GPT-4) can compensate for the gains from action knowledge, even though planning hallucinations may still occur. 
However, this benefit diminishes as task difficulty increases, and {\ours} continues to perform better on more challenging (medium and hard) test cases.
For the 70b model, the results indicate that {\ours}'s advantage is evident in lower difficulty (Easy, Medium) test cases, but this advantage decreases in Hard tests. 
Solving complex problems involves multi-step reasoning ability, where FireAct performs better in shorter in-context settings. 
The complexity of {\ours}'s setup introduces challenges in handling longer texts, leading to this observed performance trend.

\begin{table*}[t!]
\centering
\resizebox{2.05 \columnwidth}{!}{
\begin{tabular}{|c|m{1.1\linewidth}|}
\hline
\textbf{Error Type}  & \multicolumn{1}{c|}{\textbf{Results of {\ours}}} \\  
\hline 

& \textbf{Question:} Gary Harrison, began his career in the 1970s and \textcolor{darkblue_fig}{\textbf{has written over how many major-label recorded songs}} including several number-one hits, another artist who have recorded his work include Bryan White, an American country music artist? \\
& \textbf{ActionPath 1:} Start\\
& \textbf{Thought 1:} ... I need to search for information about his career and achievements.\\
& \textbf{Action 1:} Search[Gary Harrison] \\
& \textbf{Observation 1:} Gary Steven Harrison is an American songwriter. Harrison began his career in the 1970s, and \textcolor{darkblue_fig}{\textbf{has written over 300 major-label recorded songs}}, including several number one hits.\\
&\textbf{ActionPath 2:} Start->Search[Gary Harrison]\\
\textbf{Inconsistency Error} &\textbf{Thought 2:} ... I need to find out who has recorded Gary Harrison's work.\\
&\textbf{Action 2:} Search[Bryan White]\\
&\textbf{Observation 2:} Bryan Shelton White (born February 17, 1974) is an American country music singer and songwriter. Signed to Asylum Records in 1994 at age 20, White released his self-titled debut album that year.\\
&\textbf{ActionPath 3:} Start->Search[Gary Harrison]->Search[Bryan White]\\
&\textbf{Thought 3:} ...\textbf{\textcolor{darkblue_fig}{Bryan White has recorded Gary Harrison's work} \textcolor{red_fig}{(Inconsistency).}} No further actions are needed as the necessary information has been obtained. \\
&\textbf{Action 3:} \textcolor{green_fig}{\textbf{Finish[Bryan White]}}\\
&\textbf{Observation 3:} \textcolor{red_fig}{\textbf{Answer is INCORRECT}}\\
&\textbf{Correct Answer:} \textbf{\textcolor{green_fig}{\textbf{300}}} \\
\hline 

& \textbf{Question:} Are Ruggero Deodato from Italy, and Mexican Alejandro Springall, both film directors? \\
&\textbf{ActionPath 1:} Start\\
&\textbf{Thought 1:} ... To determine if Ruggero Deodato from Italy and Mexican Alejandro Springall are both film directors, it's essential to research their professional backgrounds and achievements. A ""Search"" action seems appropriate to gather this information about each individual.\\
&\textbf{Action 1:} Search[Ruggero Deodato]\\
&\textbf{Observation 1:} \textbf{\textcolor{darkblue_fig}{Ruggero Deodato (7 May 1939 – 29 December 2022) was an Italian film director}}, screenwriter, and actor...\\
&\textbf{ActionPath 2:} Start->Search[Ruggero Deodato]\\
\textbf{Summarization Error} &\textbf{Thought 2:} ...  Ruggero Deodato is an Italian film director. I need to search Alejandro Springall next and find his profession.\\
&\textbf{Action 2:} Search[Alejandro Springall]\\
&\textbf{Observation 2:} \textbf{\textcolor{darkblue_fig}{Alejandro Springall is a Mexican film director and producer.}} Springall studied filmmaking at the London Film School. He returned to Mexico City in 1991 and started working with Mexican film producer Bertha Navarro, from whom he learned most of his producing skills...\\
&\textbf{ActionPath 3:} Start->Search[Ruggero Deodato]->Search[Alejandro Springall]\\
&\textbf{Thought 3:} ... \textbf{\textcolor{darkblue_fig}{Alejandro Springall is a film producer. So Ruggero Deodato and Alejandro Springall are not both film directors}}\textcolor{red_fig}{~\textbf{(Summarization).}}
No further actions are needed as the necessary information has been obtained.\\
&\textbf{Action 3:} \textcolor{green_fig}{\textbf{Finish[no]}}\\
&\textbf{Observation 3:} \textcolor{red_fig}{\textbf{Answer is INCORRECT}}\\
&\textbf{Correct Answer:} \textcolor{green_fig}{\textbf{yes}}  \\
\hline
\end{tabular}}
\caption{Error Cases in HotpotQA.}
\label{tab:error_analysis_cases}
\end{table*}

\section{Prompt}
\label{app:prompt}
\subsection{Prompt Format}
\begin{figure*}
    \centering
    \includegraphics[width=1\textwidth]{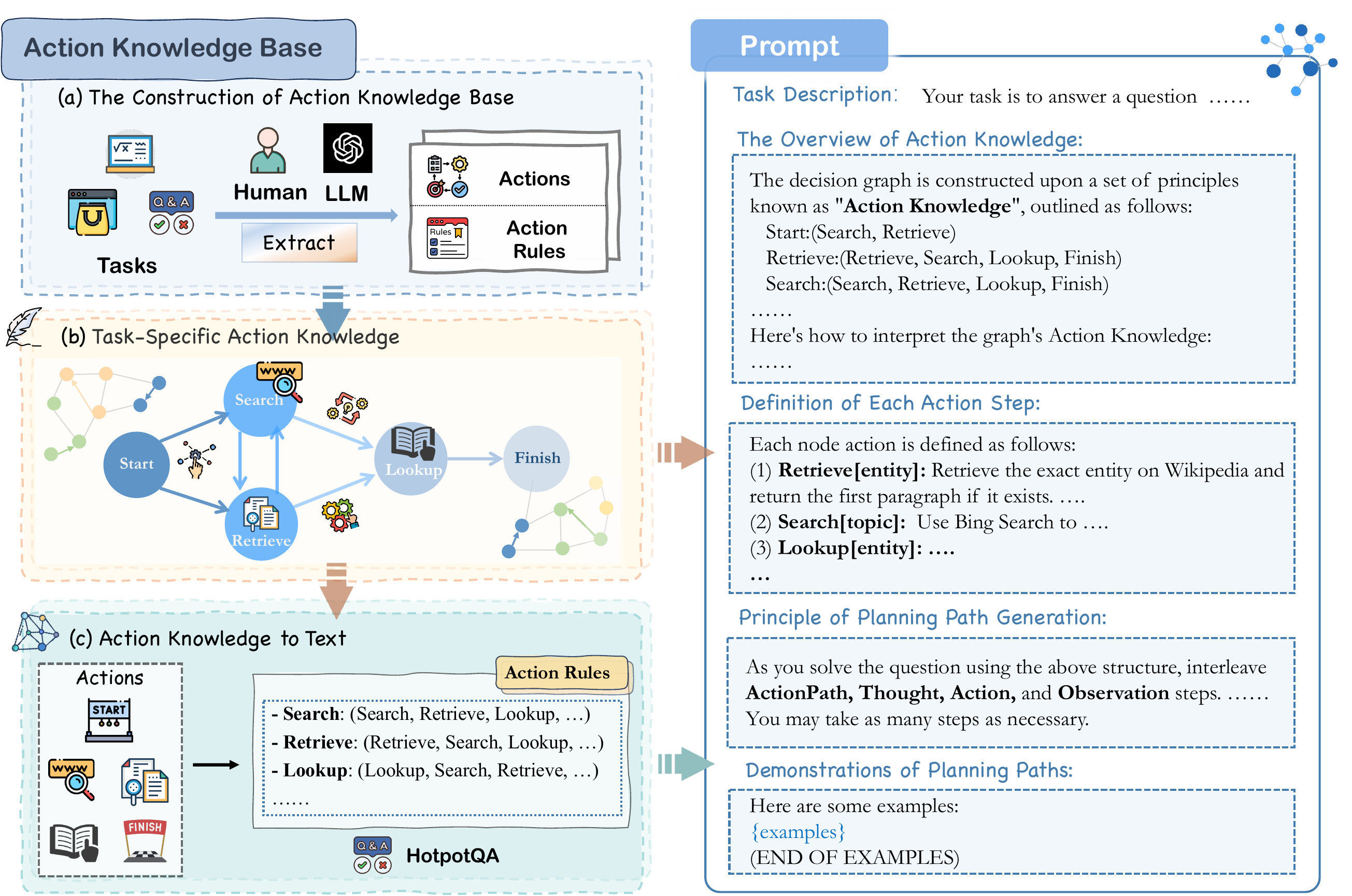}
    \caption{The Path Generation process of {\ours}.}
    \label{fig:prompt}
\end{figure*}

See Figure~\ref{fig:prompt}.

\subsection{Prompt for HotpotQA}
\label{app:prompt_hotpotqa}
See Table~\ref{tab:prompt_hotpotqa}.
\begin{table*}[t!]
    \centering
    \renewcommand\arraystretch{1}
    \scalebox{0.95}{
    \begin{tabular}{p{15cm}}
    \toprule
    \large{\textbf{Prompt for HotpotQA}}\\
    \toprule
    Your task is to answer a question using a specific graph-based method. You must navigate from the "Start" node to the "Finish" node by following the paths outlined in the graph. The correct path is a series of actions that will lead you to the answer.\par
The decision graph is constructed upon a set of principles known as "Action Knowledge", outlined as follows:\par
    \qquad Start:(Search, Retrieve)\par
    \qquad Retrieve:(Retrieve, Search, Lookup, Finish)\par
   \qquad Search:(Search, Retrieve, Lookup, Finish)\par\par
   \qquad Lookup:(Lookup, Search, Retrieve, Finish)\par
   \qquad Finish:()\par
Here's how to interpret the graph's Action Knowledge:\par
 \qquad From "Start", you can initiate with either a "Search" or a "Retrieve" action.\par
 \qquad At the "Retrieve" node, you have the options to persist with "Retrieve", shift to "Search", experiment with "Lookup", or advance to "Finish". \par
 \qquad At the "Search" node, you can repeat "Search", switch to "Retrieve" or "Lookup", or proceed to "Finish".\par
  \qquad At the "Lookup" node, you have the choice to keep using "Lookup", switch to "Search" or "Retrieve", or complete the task by going to "Finish".\par
  \qquad The "Finish" node is the final action where you provide the answer and the task is completed.\par
Each node action is defined as follows:\par
 \qquad(1) Retrieve[entity]: Retrieve the exact entity on Wikipedia and return the first paragraph if it exists. If not, return some similar entities for searching. \par
 \qquad(2) Search[topic]:  Use Bing Search to find relevant information on a specified topic, question, or term.\par
 \qquad(3) Lookup[keyword]: Return the next sentence that contains the keyword in the last passage successfully found by Search or Retrieve.\par  
 \qquad(4) Finish[answer]: Return the answer and conclude the task.\par  
As you solve the question using the above graph structure, interleave ActionPath, Thought, Action, and Observation steps. ActionPath documents the sequence of nodes you have traversed within the graph. Thought analyzes the current node to reveal potential next steps and reasons for the current situation.\par 
You may take as many steps as necessary.\par
Here are some examples:\par
\{examples\}\par
(END OF EXAMPLES)\par
Question: \{question\}\{scratchpad\}\\
\bottomrule
    \end{tabular}
    }
    \caption{Prompt for HotpotQA.}
    \label{tab:prompt_hotpotqa}
\end{table*}

\subsection{Prompt for ALFWorld}
\label{app:prompt_alfworld}
See Table~\ref{tab:prompt_alfworld_pick} to Table~\ref{tab:prompt_alfworld_picktwo}.

\begin{table*}[t!]
    \centering
    \renewcommand\arraystretch{1}
    \scalebox{0.95}{
    
    \begin{tabular}{p{15cm}}
    \toprule
    \textbf{ALFWorld - Pick}  \\
   \Xhline{1pt}
Interact with a household to solve a task by following the structured "Action Knowledge". The guidelines are:\par
Goto(receptacle) -> Open(receptacle)\par
[Goto(receptacle), Open(receptacle)] -> Take(object, from: receptacle)\par
Take(object, from: receptacle) -> Goto(receptacle)\par
[Goto(receptacle), Take(object, from: receptacle)] -> Put(object, in/on: receptacle)\par\\
Here's how to interpret the Action Knowledge:\par
Before you open a receptacle, you must first go to it. This rule applies when the receptacle is closed.\par
To take an object from a receptacle, you either need to be at the receptacle's location, or if it's closed, you need to open it first. \par
Before you go to the new receptacle where the object is to be placed, you should take it. \par
Putting an object in or on a receptacle can follow either going to the location of the receptacle or after taking an object with you. 
\par\\
The actions are as follows:\par
1) go to receptacle \par
2) take object from receptacle \par
3) put object in/on receptacle \par
4) open receptacle\par\\
As you tackle the question with Action Knowledge, utilize both the ActionPath and Think steps. ActionPath records the series of actions you've taken, and the Think step understands the current situation and guides your next moves.\par\\

Here are two examples.\par
\{examples\}\par
Here is the task.
    \\
     \bottomrule
    \end{tabular}
    }
    \caption{Prompt for the \texttt{Pick} Task.}
    \label{tab:prompt_alfworld_pick}
\end{table*}

\begin{table*}[t!]
    \centering
    \renewcommand\arraystretch{1}
    \scalebox{0.95}{
    
    \begin{tabular}{p{15cm}}
         \toprule
    \textbf{ALFWorld - Light}  \\
   \Xhline{1pt}
Interact with a household to solve a task by following the structured "Action Knowledge". The guidelines are:\par
[Goto(receptacle)] -> Open(receptacle)\par
[Goto(receptacle), Open(receptacle)] -> Take(object, from: receptacle)\par
[Goto(receptacle)] -> Use(receptacle)
\par\\
Here's how to interpret the Action Knowledge:\par
Before you open a receptacle, you must first go to it. This rule applies when the receptacle is closed.\par
To take an object from a receptacle, you either need to be at the receptacle's location, or if it's closed, you need to open it first. \par
To use an receptacle, you must go to the place where it is located.
\par\\
The actions are as follows:\par
1) go to receptacle \par
2) take object from receptacle\par
3) use receptacle \par
4) open receptacle\par\\
As you tackle the question with Action Knowledge, utilize both the ActionPath and Think steps. ActionPath records the series of actions you've taken, and the Think step understands the current situation and guides your next moves.\par\\

Here are two examples.\par
\{examples\}\par
Here is the task.
    \\
     \bottomrule
    \end{tabular}
    }
    \caption{Prompt for the \texttt{Light} Task.}
    \label{tab:prompt_alfworld_light}
\end{table*}

\begin{table*}[t!]
    \centering
    \renewcommand\arraystretch{1}
    \scalebox{0.95}{
    
    \begin{tabular}{p{15cm}}
         \toprule
    \textbf{ALFWorld - Clean}  \\
   \Xhline{1pt}
Interact with a household to solve a task by following the structured "Action Knowledge". The guidelines are:\par
[Goto(receptacle)] -> Open(receptacle)\par
[Goto(receptacle), Open(receptacle)] -> Take(object, from: receptacle)\par
[Goto(receptacle), Take(object, from: receptacle)] -> Put(object, in/on: receptacle)\par
[Put(object, from: receptacle)] -> Clean(object, with: receptacle)
\par\\
Here's how to interpret the Action Knowledge:\par
Before you open a receptacle, you must first go to it. This rule applies when the receptacle is closed.\par
To take an object from a receptacle, you either need to be at the receptacle's location, or if it's closed, you need to open it first. \par
Putting an object in or on a receptacle can follow either going to the location of the receptacle or after taking an object with you. \par
To clean an object using a receptacle, the object must first be placed in or on that receptacle.
\par\\
The actions are as follows:\par
1) go to receptacle \par
2) take object from receptacle\par
3) open receptacle\par
4) put object in/on receptacle \par
5) clean object with receptacle\par\\
As you tackle the question with Action Knowledge, utilize both the ActionPath and Think steps. ActionPath records the series of actions you've taken, and the Think step understands the current situation and guides your next moves.\par\\

Here are two examples.\par
\{examples\}\par
Here is the task.
    \\
     \bottomrule
    \end{tabular}
    }
    \caption{Prompt for the \texttt{Clean} Task.}
    \label{tab:prompt_alfworld_clean}
\end{table*}

\begin{table*}[t!]
    \centering
    \renewcommand\arraystretch{1}
    \scalebox{0.95}{
    
    \begin{tabular}{p{15cm}}
         \toprule
    \textbf{ALFWorld - Heat}  \\
   \Xhline{1pt}
Interact with a household to solve a task by following the structured "Action Knowledge". The guidelines are:\par
[Goto(receptacle)] -> Open(receptacle)\par
[Goto(receptacle), Open(receptacle)] -> Take(object, from: receptacle)\par
[Goto(receptacle), Take(object, from: receptacle)] -> Put(object, in/on: receptacle)\par
[Put(object, in/on: receptacle)] -> Heat(object, with: receptacle)
\par\\
Here's how to interpret the Action Knowledge:\par
Before you open a receptacle, you must first go to it. This rule applies when the receptacle is closed.\par
To take an object from a receptacle, you either need to be at the receptacle's location, or if it's closed, you need to open it first. \par
Putting an object in or on a receptacle can follow either going to the location of the receptacle or after taking an object with you. \par
To heat an object using a receptacle, the object must first be placed in or on that receptacle.
\par\\
The actions are as follows:\par
1) go to receptacle \par
2) take object from receptacle\par
3) open receptacle\par
4) put object in/on receptacle \par
5) heat object with receptacle\par\\
As you tackle the question with Action Knowledge, utilize both the ActionPath and Think steps. ActionPath records the series of actions you've taken, and the Think step understands the current situation and guides your next moves.\par\\

Here are two examples.\par
\{examples\}\par
Here is the task.
    \\
     \bottomrule
    \end{tabular}
    }
    \caption{Prompt for the \texttt{Heat} Task.}
    \label{tab:prompt_alfworld_heat}
\end{table*}

\begin{table*}[t!]
    \centering
    \renewcommand\arraystretch{1}
    \scalebox{0.95}{
    
    \begin{tabular}{p{15cm}}
         \toprule
    \textbf{ALFWorld - Cool}  \\
   \Xhline{1pt}
Interact with a household to solve a task by following the structured "Action Knowledge". The guidelines are:\par
[Goto(receptacle)] -> Open(receptacle)\par 
[Goto(receptacle), Open(receptacle)] -> Take(object, from: receptacle)\par
[Goto(receptacle), Take(object, from: receptacle)] -> Put(object, in/on: receptacle)\par
[Put(object, in/on: receptacle)] -> Cool(object, with: receptacle)
\par\\
Here's how to interpret the Action Knowledge:\par
Before you open a receptacle, you must first go to it. This rule applies when the receptacle is closed.\par
To take an object from a receptacle, you either need to be at the receptacle's location, or if it's closed, you need to open it first. \par
Putting an object in or on a receptacle can follow either going to the location of the receptacle or after taking an object with you. \par
To cool an object using a receptacle, the object must first be placed in or on that receptacle.
\par\\
The actions are as follows:\par
1) go to receptacle \par
2) take object from receptacle\par
3) open receptacle\par
4) put object in/on receptacle \par
5) cool object with receptacle\par\\
As you tackle the question with Action Knowledge, utilize both the ActionPath and Think steps. ActionPath records the series of actions you've taken, and the Think step understands the current situation and guides your next moves.\par\\

Here are two examples.\par
\{examples\}\par
Here is the task.
    \\
     \bottomrule
    \end{tabular}
    }
    \caption{Prompt for the \texttt{Cool} Task.}
    \label{tab:prompt_alfworld_cool}
\end{table*}

\begin{table*}[t!]
    \centering
    \renewcommand\arraystretch{1}
    \scalebox{0.95}{
    
    \begin{tabular}{p{15cm}}
         \toprule
    \textbf{ALFWorld - Pick Two}  \\
   \Xhline{1pt}
   Interact with a household to solve a task by following the structured "Action Knowledge". The guidelines are:\par
Goto(receptacle) -> Open(receptacle)\par
[Goto(receptacle), Open(receptacle)] -> Take(object, from: receptacle)\par
Take(object, from: receptacle) -> Goto(receptacle)\par
[Goto(receptacle), Take(object, from: receptacle)] -> Put(object, in/on: receptacle)
\par\\
Here's how to interpret the Action Knowledge:\par
Before you open a receptacle, you must first go to it. This rule applies when the receptacle is closed.\par
To take an object from a receptacle, you either need to be at the receptacle's location, or if it's closed, you need to open it first. \par
Before you go to the new receptacle where the object is to be placed, you should take it. \par
Putting an object in or on a receptacle can follow either going to the location of the receptacle or after taking an object with you. \par
Ensure the first object is placed before proceeding to deposit the second object.
\par\\
The actions are as follows:\par
1) go to receptacle \par
2) take object from receptacle \par
3) put object in/on receptacle \par
4) open receptacle\par\\
As you tackle the question with Action Knowledge, utilize both the ActionPath and Think steps. ActionPath records the series of actions you've taken, and the Think step understands the current situation and guides your next moves.\par\\

Here are two examples.\par
\{examples\}\par
Here is the task.
    \\
     \bottomrule
    \end{tabular}
    }
    \caption{Prompt for the \texttt{Pick Two} Task.}
    \label{tab:prompt_alfworld_picktwo}
\end{table*}

\end{document}